\documentclass[lettersize,journal]{IEEEtran}

\usepackage{amsmath,amsfonts}
\usepackage{amsthm}
\usepackage{array}
\usepackage{booktabs}

\usepackage[table,xcdraw,dvipsnames]{xcolor}
\usepackage{colortbl}

\usepackage{graphicx}
\usepackage{tikz}
\usepackage{pgfplots}
\usepackage{subcaption}

\usepackage{adjustbox}
\usepackage{multirow}
\usepackage{multicol}
\usepackage{makecell}
\usepackage{tabularx}
\usepackage{float}
\usepackage{placeins}
\usepackage{stfloats}
\usepackage{lscape}
\usepackage{afterpage}

\usepackage{bm}
\usepackage{algorithm}
\usepackage{algorithmic}
\usepackage{fontawesome5}
\usepackage{pifont}
\usepackage{textcomp}

\usepackage{enumitem}

\usepackage{xspace}
\usepackage{url}
\usepackage{verbatim}

\usepackage{cite}
\usepackage{xr}

\usepackage[colorlinks=true,
            linkcolor=blue,
            citecolor=blue,
            urlcolor=blue]{hyperref}

\usepackage{cleveref}

\newcommand{\eg}{\emph{e.g.}\xspace}
\newcommand{\ie}{\emph{i.e.}\xspace}

\definecolor{ForestGreen}{RGB}{34, 139, 34}
\definecolor{BrickRed}{RGB}{178, 34, 34}
\definecolor{NavyBlue}{RGB}{0, 100, 200}

\theoremstyle{plain}
\newtheorem{theorem}{Theorem}[section]

\theoremstyle{definition}

\theoremstyle{remark}

\newcommand{\efthygeo}[1]{\textcolor{ForestGreen}{ #1}}

\newcommand{\Gauss}{K^{\text{gauss}}}
\newcommand{\gauss}{\kappa^{\text{gauss}}}
\newcommand{\riesz}{\kappa^{\text{riesz}}}
\newcommand{\Logar}{K^{\text{log}}}
\newcommand{\logar}{\kappa^{\text{log}}}

\newcommand{\Linfoncea}{L_{\textnormal{InfoNCE}}}

\newcommand{\Lmvdhel}{L_{\textnormal{MV-DHEL}}}
\newcommand{\Lmvinfonce}{L_{\textnormal{MV-InfoNCE}}}
\newcommand{\Lpvc}{L_{\textnormal{PVC}}}
\newcommand{\Lpwe}{L_{\textnormal{pwe}}}
\newcommand{\Lavg}{L_{\textnormal{avg}}}
\newcommand{\Lpair}{L_{\textnormal{pair}}}

\newcommand{\funsimple}{f}

\newcommand{\pdata}{p}

\DeclareMathOperator{\E}{\mathbb{E}}

\usepackage{mathtools}

\renewcommand{\b}[1]{\mathbf{#1}}

\newcommand{\bU}{\b{U}}

\newcommand{\bX}{\b{X}}
\newcommand{\bY}{\b{Y}}

\newcommand{\bu}{\b{u}}
\newcommand{\bv}{\b{v}}
\newcommand{\bx}{\b{x}}
\newcommand{\by}{\b{y}}

\newcommand{\cN}{\mathcal{N}}
\newcommand{\cP}{\mathcal{P}}

\newcommand{\cT}{\mathcal{T}}
\newcommand{\cU}{\mathcal{U}}

\newcommand{\cX}{\mathcal{X}}

\newcommand{\cZ}{\mathcal{Z}}

\DeclareMathAlphabet{\mathsfit}{\encodingdefault}{\sfdefault}{m}{sl}
\SetMathAlphabet{\mathsfit}{bold}{\encodingdefault}{\sfdefault}{bx}{n}
\newcommand{\tens}[1]{\bm{\mathsfit{#1}}}
\newcommand{\tU}{\tens{U}}
\newcommand{\tX}{\tens{X}}

\usepackage{mathtools}

\begin{document}

\title{A Principled Framework for Multi-View \\ Contrastive Learning}

\author{Panagiotis Koromilas, Efthymios Georgiou, Giorgos Bouritsas, Theodoros Giannakopoulos, Mihalis A. Nicolaou, and Yannis Panagakis
\thanks{P. Koromilas, G. Bouritsas and Y. Panagakis are with the Department of Informatics and Telecommunications, National and Kapodistrian University of Athens.}%
\thanks{G. Bouritsas and Y. Panagakis are also with Archimedes AI/Athena Research Center.}%
\thanks{E. Georgiou is with ILSP/Athena Research Center }
\thanks{T. Giannakopoulos is with NCSR ``Demokritos".}%
\thanks{M. A. Nicolaou is with The Cyprus Institute.}%
}

\maketitle

\begin{abstract}
Contrastive Learning (CL), a leading paradigm in Self-Supervised Learning (SSL), typically relies on pairs of data views generated through augmentation. While multiple augmentations per instance (more than two) improve generalization in supervised learning, current CL methods handle additional views suboptimally by simply aggregating different pairwise objectives. This approach suffers from four critical limitations: (L1) it utilizes multiple optimization terms per data point resulting to conflicting objectives, (L2) it fails to model all interactions across views and data points, (L3) it inherits fundamental limitations (\eg alignment-uniformity coupling) from pairwise CL losses, and (L4) it prevents fully realizing the benefits of increased view multiplicity observed in supervised settings. We address these limitations through two novel loss functions: MV-InfoNCE, which extends InfoNCE to incorporate all possible view interactions simultaneously in one term per data point, and MV-DHEL, which decouples alignment from uniformity across views while scaling interaction complexity with view multiplicity. Both approaches are theoretically grounded - we prove they asymptotically optimize for alignment of all views and uniformity, providing principled extensions to multi-view contrastive learning. Our empirical results on ImageNet1K and three other datasets demonstrate that our methods consistently outperform existing multi-view approaches and effectively scale with increasing view multiplicity. We also apply our objectives to multimodal data and show that, in contrast to other contrastive objectives, they can scale beyond just two modalities. Most significantly, ablation studies reveal that MV-DHEL with five or more views effectively mitigates dimensionality collapse by fully utilizing the embedding space, thereby delivering multi-view benefits observed in supervised learning. \\
Code: \href{https://github.com/pakoromilas/Multi-View-CL.git}{https://github.com/pakoromilas/Multi-View-CL.git}
\end{abstract}

\begin{IEEEkeywords}
Contrastive learning, multi-view contrastive learning, multimodal contrastive learning, self-supervised learning.
\end{IEEEkeywords}

\begin{figure*}
    \centering
    \includegraphics[width=0.8\textwidth]{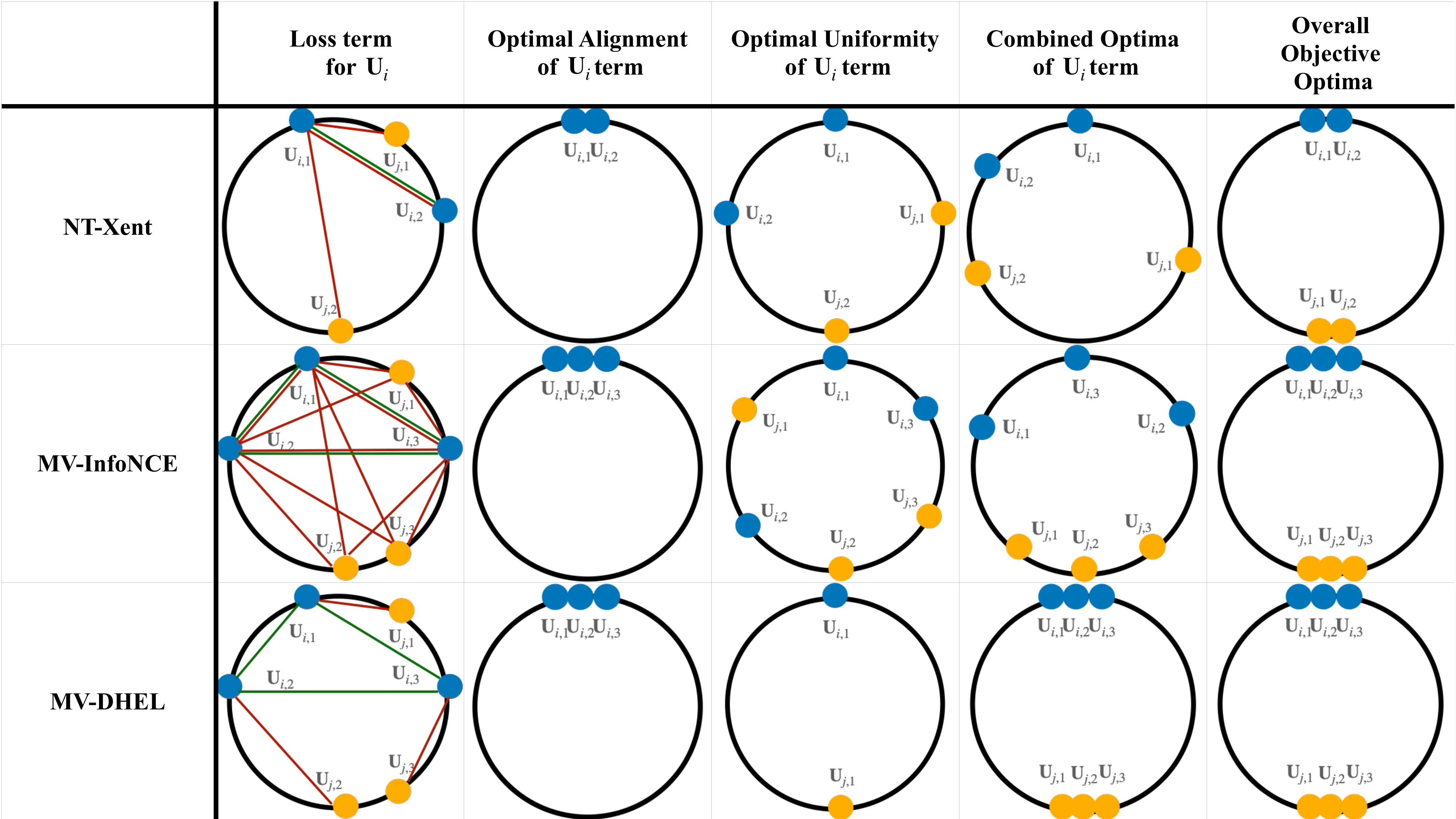}
\caption{\textbf{Alignment and uniformity optimisation in different objectives}. This figure illustrates how three CL methods, NT-Xent, our MV-InfoNCE, and our MV-DHEL, optimise data representations. Given a representation $\bU_i$ with different views $\bU_{i,l}$, we visualize optimal interactions only for the loss term associated with $\bU_i$. \textcolor{ForestGreen}{Green lines} indicate positive interactions (alignment), while \textcolor{BrickRed}{red lines} represent negative interactions (uniformity). Each column shows a different optimisation component. The figure demonstrates two key advantages of our approaches: (1) \textbf{both our methods leverage all multi-view interactions} within a single objective, unlike NT-Xent which only considers pairwise interactions, and (2) in our \textit{MV-DHEL}, alignment and uniformity terms remain fully \textbf{disentangled}, ensuring that \textit{each loss term is optimized for the same point configuration as the overall objective}, making it \textbf{easier to reach the global optima}.}
\label{fig:coupling} 
\end{figure*}

\section{Introduction}
\label{sec:intro}

\IEEEPARstart{S}{elf-Supervised Learning} (SSL) enables learning representations from unlabeled data by exploiting inherent data structure and invariances. Among SSL approaches, Contrastive Learning (CL) has emerged as a leading paradigm by optimizing two complementary objectives: maximizing similarity (alignment) between different views of the same instance while ensuring varying instances remain distinguishable (uniformity / energy) \cite{wang2020understanding}. In this context, \textit{data views refer to variations of the same data point}, such as an image captured from different angles or under different lighting conditions, which can occur naturally in the data or be systematically created \textit{through augmentation} techniques \cite{philip2019learning}.

The benefits of employing multiple views in learning extend beyond mere data augmentation. In supervised representation learning, multiple views per data point improve learning in three key ways: (i) enabling higher learning rates through more stable gradient updates \cite{fort2021drawing}, (ii) accelerating convergence by reducing gradient variance \cite{hoffer2020augment}, and (iii) improving out-of-distribution generalization through implicit regularization \cite{lin2024good}. Recognizing these advantages, several SSL methods have adopted multiple views: SwAV \cite{caron2020unsupervised} employs multi-resolution crops for clustering-based learning, DINO \cite{caron2021emerging} combines global and local views for knowledge distillation, and VICRegL \cite{bardes2022vicregl} enforces multi-scale consistency in representation learning. Each method demonstrates that incorporating \textit{more than two views} improves representation quality and downstream performance.

Current multi-view CL methods solely aggregate pairwise losses \cite{tian2020contrastive, shidani2024polyview}, an approach fundamentally limited in four critical aspects. \textbf{Multi-term optimization (L1)}: In current objectives increasing view multiplicity also increases the number of loss terms per instance (one for each view), forcing each representation to satisfy multiple potentially conflicting objectives. \textbf{Missing concurrent interactions across all views (L2)}: Although similarity measures are employed to guide optimization across views and instances, current objectives fail to capture all possible interactions among views within a batch thus not guaranting optimization of the desideratum which requires simultaneously aligning all n-views, and contrasting each single-view datapoint to all views of the remaining datapoints in the batch. \textbf{Alignment-uniformity coupling (L3)}: Each pairwise comparison inherits fundamental CL issues where view interactions contribute to both alignment and uniformity calculations, resulting in conflicting objectives \cite{koromilas2024bridging} that worsen as view multiplicity increases. \textbf{Limited transfer of multi-view benefits (L4)}: Unlike the benefits observed in supervised learning when employing multiple views \cite{hoffer2020augment, fort2021drawing}, simply increasing view multiplicity by aggregating pairwise losses does not capture the desired interactions across multiple views that preserve the intrinsic dimensionality of representations. These limitations prevent current methods from fully leveraging the potential benefits of multiple views.

We expect that a properly designed multi-view contrastive learning framework should realize two key benefits that current methods fail to achieve. \textbf{Better optimization}: Multiple views per batch should (i) provide a more accurate approximation of the true objective, whose optimal value is attained with perfect uniformity \cite{wang2020understanding} thus \textit{improving uniformity}, and (ii) increase the algorithm's capability to achieve desired invariances by simultaneously aligning all views thus leading to \textit{stronger alignment}. \textbf{Mitigating dimensionality collapse}: Drawing from supervised learning, multiple augmented views, when properly leveraged, should reduce gradient variance and preserve the intrinsic dimensionality of representations \cite{hoffer2020augment, fort2021drawing}, addressing the problem where learned representations utilize only a small fraction of available embedding dimensions \cite{jing2022understanding}.

To achieve these benefits and address the limitations of current methods, we introduce a framework to express each multi-view contrastive loss and identify three fundamental principles for a theoretically sound objective with the desired behavior. First, given a data point $i$ and a view of interest $l$, \textcolor{ForestGreen}{(P1) simultaneous alignment} requires all other views of the same data point to be simultaneously aligned, ensuring invariance to all transformations without competing objectives. Second, \textcolor{BrickRed}{(P2) accurate energy term} mandates that the uniformity component must capture all pairwise interactions in the representation configuration. Third, \textcolor{NavyBlue}{(P3) one term per data instance} maintains a single optimization term per instance for the complete objective, ensuring better optimization. Current objectives violate all three principles, resulting in the four limitations (L1-4) identified above.

Guided by these principles, we introduce two novel multi-view contrastive objectives that address the fundamental limitations of existing approaches:
\begin{enumerate}
    \item \textbf{MV-InfoNCE}: We generalize InfoNCE to capture interactions between all views simultaneously, rather than just summing pair-wise losses, in a single term adressing both (L1) and (L2). 
    \item \textbf{MV-DHEL}: We extend the DHEL \cite{koromilas2024bridging} loss to decouple alignment from uniformity across views and enable richer interactions that scale with the number of views. This resolves the alignment-uniformity coupling (L3) that becomes more severe as view multiplicity increases.
\end{enumerate}

We theoretically prove that both objectives share the same minima as traditional two-view contrastive learning \cite{wang2020understanding}, providing \textit{mathematically sound extensions} of CL to the multi-view context. Empirically, our methods demonstrate three key advantages. First, they achieve \textit{higher downstream accuracy} scores across multiple datasets. Second, they show improved \textit{scalability with increasing number of views}. Third, they effectively \textit{mitigate dimensionality collapse}, with MV-DHEL preserving the intrinsic dimensionality of representations when using five or more views (addressing limitation L4). Finally, although designed for single-view unimodal learning, we demonstrate that, unlike existing contrastive methods \cite{radford2021learning}, which struggle to generalize beyond two modalities \cite{ruan2024tricolo, liu2021contrastive, sun2024contextual}, our approach is \textit{effective in multimodal settings} that extend beyond the typical two-modality framework \cite{radford2021learning}.

Our work addresses fundamental limitations in current multi-view CL with the following key contributions:

\begin{enumerate}
    \item \textbf{Principled Multi-View Formulation}: We develop a mathematical framework that properly extends CL from pairwise to multi-view settings, enabling the modeling of simultaneous interactions between all n-views in a single term rather than simply aggregating multi-term pairwise comparisons.
    \item \textbf{Novel Loss Functions}: We introduce (i) \textbf{MV-InfoNCE}: a natural extension of InfoNCE that incorporates all possible view interactions in a generalized objective, resolving the limitations of multi-term optimization per data point (L1) and that of missing concurrent interactions (L2), and (ii) \textbf{MV-DHEL} a multi-view loss function that decouples alignment from uniformity across views, addressing the coupling problems (L3).
    \item \textbf{Theoretical Guarantees}: We prove that both proposed objectives share the same asymptotic behavior as the traditional InfoNCE loss, establishing them as \textit{theoretically sound extensions} to the multi-view setting.
    \item \textbf{Empirical Advances}: Through extensive experiments on four datasets (including ImageNet1K), we demonstrate that: (i) our methods \textit{consistently outperform} existing multi-view contrastive approaches, (ii) performance \textit{effectively scales} with increasing view multiplicity, (iii) MV-DHEL preserves the benefits of multiple augmentations observed in supervised learning (L4), and with sufficient views, \textit{mitigates dimensionality collapse}
    \item \textbf{Multimodal Applicability}: Unlike existing contrastive methods predominantly designed for bimodal settings, our framework is directly applicable to learning across three or more modalities, as validated by our experiments on multimodal sentiment analysis.
\end{enumerate}

These contributions collectively advance the state-of-the-art in self-supervised learning by \textit{addressing longstanding limitations} of contrastive approaches and providing both theoretical foundations and practical implementations for principled multi-view learning paradigms.

\section{Related Work}
\label{sec:related_work}
\textbf{Contrastive Learning.} Contrastive learning was first introduced in \cite{chopra2005learning} and later generalised to the (N+1) tuple loss \cite{sohn2016improved}, culminating in the widely adopted InfoNCE loss used in contrastive predictive coding \cite{oord2018representation}. The NT-Xent variant \cite{chen2020simple}, which normalises the temperature in the InfoNCE loss, along with techniques such as projection head, sampled augmentations and large batch sizes, has become foundational in contemporary contrastive learning methods \cite{chen2020simple, dwibedi2021little, yeh2022decoupled}. Despite its effectiveness, InfoNCE has notable limitations; performance improves with an increased number of negative samples, which necessitates large batch sizes and strategies like hard-negative sampling \cite{chen2020simple, tian2020contrastive, he2020momentum, RobinsonCSJ21}. Additionally, learned representations often underutilise the available dimensions, leading to dimensionality collapse \cite{hua2021feature, JingVLT22}. In theory, InfoNCE aims to optimise for asymptotically aligned and uniformly distributed representations \cite{wang2020understanding}. Recent generalisations have extended this understanding to broader instances of InfoNCE and kernel-based losses \cite{koromilas2024bridging}. In this work, we propose two objectives that utilise multiple views and show that in theory, they exhibit the same asymptotic behaviour to InfoNCE and in practice, they effectively improve performance while mitigating dimensionality collapse.

\textbf{Augmentation multiplicity in Supervised Learning.} Several studies highlight the positive impact of increased augmentation multiplicity \cite{philip2019learning} on model performance. Drawing multiple augmented samples per image has been shown to reduce gradient variance, stabilizing training and enabling a larger learning rate, thereby improving performance per training step \cite{fort2021drawing}. Further evidence indicates that diverse data augmentations, even if inconsistent with the data distribution, can enhance performance in out-of-distribution (OOD) scenarios, sometimes outperforming additional training data \cite{geiping2023how}. Generalisation is further positively impacted by data augmentation through the implicit introduction of spectral regularisation \cite{lin2024good}. The concept of augmentation multiplicity, as explored by \cite{hoffer2020augment, fort2021drawing}, reveals a key insight: by drawing multiple augmented samples of each unique image within a batch, one can retain the beneficial bias introduced by data augmentation while suppressing gradient variance. Additionally, the learned invariances introduced by data augmentation have been quantified, showing that popular random-resized-crop (RRC) augmentation effectively combines translation and scaling \cite{bouchacourt2021grounding}. 

\textbf{Multiple views in SSL.} The conventional approach in SSL, which used paired views to address the lack of labels, has shifted toward using multiple views of the same data. Leading methods like SwAV \cite{caron2020unsupervised}, DINO \cite{caron2021emerging}, and VICRegL \cite{bardes2022vicregl} exemplify this shift by integrating more than two views, resulting in improved performance across various SSL tasks. For instance, SwAV employs a multi-crop strategy to extract information from multiple random crops of different sizes, enabling richer representations. Similarly, DINO uses a combination of two global views along with several local views to leverage local-to-global relationships, while VICRegL \cite{bardes2022vicregl} promotes learning across multiple scales for enhanced generalisation. EMP-SSL \cite{tong2023empssl} leverages multiple views to reduce the training epochs needed for convergence, and Whitening MSE Loss \cite{ermolov2021whitening} uses the same, among others, technique to prevent collapse in non-contrastive (positive-only) SSL, without the need for momentum networks and stop-gradient operations.

\textbf{Multi-View Contrastive Learning.} The effectiveness of leveraging multiple views in CL has been demonstrated by aggregating pairwise losses from views generated through various means, such as multiple projection heads \cite{wang2024adaptive}, image scales \cite{li2023global}, classes \cite{shah2023multi}, and feature levels \cite{xu2022multi}. However, the way to efficiently utilise multiple views in Contrastive Learning has been indipendently studied. 
 An early approach by \cite{tian2020contrastive} framed multi-view learning as a collection of independent tasks, treating each combination of views separately and simply aggregating pairwise losses. Recently, \cite{shidani2024polyview} has incorporated another aggregation method, while utilizing optimal transport on a high similarity tensor was proposed by \cite{piran2024contrasting}. However the former does not model all direct interactions across views while it exhibits alignment and uniformity coupling that hurts optimisation. The latter is based on optimal transport and thus its theoretical behavior and connection to the optima of contrastive learning remains undefined. In this work we propose multi-view contrastive learning objectives that (i) model all direct interactions across views, (ii) alleviate alignment-uniformity coupling, and (iii) exhibit the same asymptotic behaviour as two-view CL.

\section{Preliminaries on Contrastive Learning}\label{sec:prelims}
\subsection{Notation}
Vectors and matrices are denoted by lowercase and uppercase  bold letters respectively, $\bu$, $\bU$, tensors are represented by uppercase bold upright letters $\tU$ and sets with calligraphic letters $\cU$. An element (scalar) within a matrix/tensor $\tU$ is accessed using subscript notation, such as $\tU_{i,j,k}$. Fibers (generalisation of rows and columns from matrices to tensors) are represented by fixing all indices except one. For instance, mode-1 fibers of $\tU$ are denoted by $\tU_{:,j,k}$. Similarly, slices (matrices) of a tensor are formed by fixing one index, \ie $\tU_{i,:,:}$. 
To denote vertical (row-wise) concatenation of matrices $\bX$ and $\bY$, we use
$[\bX;\bY]$, while for depth stacking, i.e., combining matrices as slices of a tensor along a new dimension, we use $[\bX,\bY]$. Further, we denote with $[N]$ the set of indices $\{1,...,N\}$ with cardinality N.

\subsection{Contrastive Learning Setup}

\textit{Contrastive Learning (CL)} is a paradigm for learning data representations without having access to labels, but based solely on information about similarities between inputs, or more strictly speaking, about \textit{downstream task invariances}. 
 
Formally, let $\cX$ be the \textit{input space}, where the data points reside
and $\cZ$ the \textit{embedding space}. Additionally, denote the (unknown) underlying data distribution on $\cX$ with $p_{\text{init}}$. Consider an \textit{encoder} $f_{\boldsymbol{\theta}}:\cX \to \cZ$, such as a neural network, parametrised by $\boldsymbol{\theta} \in \Theta$, which maps input data points to their corresponding representations. In this setup, we assume $\cZ = \mathbb{S}^{d-1} = 
\{\bu \in \mathbb{R}^d \mid \|\bu\| = 1\}$, the unit sphere. We use  $\bx$ to represent a specific view of input data points, and $\bu$ to denote the corresponding representations. 

In the \textbf{multi-view learning setup}, each data point is represented by a collection of $N$ different views, $\bX = [\bx_1; \dots ; \bx_N]$ and
$\tX = [\bX_1, \dots, \bX_M] $ denotes a collection (batch) of $M$ input data points. The corresponding set of $M$ representations is given by $\tU = [\bU_1, \dots, \bU_M] \in \mathbb{R}^{M \times N \times d}$, where $\bU$ equals $f_{\boldsymbol{\theta}}(\bX) = [f_{\boldsymbol{\theta}}(\bx_1); \dots; f_{\boldsymbol{\theta}}(\bx_M)]$.  
In conventional CL, $N=2$ and the encoder is trained by optimising an objective that encourages the representations of these two views (\textit{positive pairs}) to be close in $\cZ$ and those of the rest of the datapoints in the $M$-collection (\textit{negatives}) to be further. Similarly, in Multi-View CL this objective will be extended so as to encourage all views of the same data point to be close in the embedding space.

\textbf{Sampling Process}: We collect multi-view datapoints as follows: First, we sample a data point $\bx_{\text{init}} \in \cX$ from the \textit{initial distribution} $p_{\text{init}}$ on $\cX$ (\ie\ the one from which we sample the data points in our dataset) and subsequently we independently sample $N$ \textit{transformation operators} ${T_i: \cX \to \cX}$ from a known distribution $p_{T}$ on a space of available transformations $\cT$. \textit{The transformation operators encode the symmetries of the data, \ie\ it is expected that the downstream tasks will be invariant to them.} The resulting N-view datapoint are tuples of the form $[\bx_1; \dots ; \bx_N] = \left(T_1\left(\bx_{\text{init}}\right),..., T_N\left(\bx_{\text{init}}\right)\right)$.

\subsection{A Common Framework for Pairwise Losses}

The objective of CL is to optimise an expected value, often represented by functions within the InfoNCE family (\eg NT-Xent \cite{chen2020simple} and DHEL \cite{koromilas2024bridging}). Using tensor notation, \textit{mini-batch losses} (expected value estimators) can be expressed as follows:
\begin{equation}\label{eq:mb_loss}
\begin{split}
    &L_{\text{CL}}(\tU ; l)  =
     -\frac{1}{M}  \sum_{
    i\in[M]} \log\left(\frac{e^{\tU_{i, {l},:}^{\top} \tU_{i,\textcolor{ForestGreen}{l'},:}{/\tau}}}
{\sum_{\textcolor{BrickRed}{(j,m)\in \cN(i,l)}}e^{{\tU_{i,l,:}^{\top} \textcolor{BrickRed}{\tU_{j, m,:}}}{/\tau}}}\right) 
\end{split}
\end{equation}

where $l \in [2]$ is the view of interest and $\textcolor{ForestGreen}{l' = (2-l) \mod 2}$, i.e. the additional view (\textbf{positive pair}) of $l$ (positives are color-coded with \textcolor{ForestGreen}{green}) and $\textcolor{BrickRed}{\cN(i,l)}$ is the \textbf{negative index set} (negatives are color-coded with \textcolor{BrickRed}{red}), i.e. the set of indices that determines which datapoints should be contrasted to the datapoint indexed by $(i,l)$. For example, we obtain NT-Xent by setting $\textcolor{BrickRed}{\cN(i,l)} = \{(j,m) \mid j \in [M], m = [2]\}$ and DHEL by setting 
$\textcolor{BrickRed}{\cN(i,l)} = \{(j,m) \mid j \in [M], j \neq i, m = l\}$.

We further introduce a convenient alternative formulation that will be useful to roll out our methodology in \Cref{sec:method}. Every CL loss can be decomposed in two terms that represent \textit{alignment} and \textit{uniformity}.\footnote{Conceptually, alignment measures the similarity among views of the same data point and uniformity measures the distance among views of different data points in the dataset \cite{wang2020understanding}.}
To make our formulation compact, we draw inspiration from \cite{koromilas2024bridging} and for each pair of single-view datapoints we will be using \textit{kernel} notation:
\begin{equation}\label{eq:kernel_sums}
K\left(\bu, \bv\right) =  
\kappa(\Vert \bu - \bv\Vert^2_2),
\end{equation}
where $\kappa$ is a single-input scalar function 
({see \Cref{sec:ad_notation} for the exact requirements}) and $\bu, \bv$ are single-view datapoints. All InfoNCE variants use the gaussian kernel
 $
\kappa(\bx) = e^{\frac{2 - \bx}{2\tau}}$.
Using \Cref{eq:kernel_sums}, we reformulate \Cref{eq:mb_loss} as follows:
\begin{equation}\label{eq:align_unif}
\begin{split}
{L_{\text{CL}}}(\tU ; l) &= \frac{-1}{M}\sum_{
i \in [M]}
\log  K(\tU_{i,
{l},:}, \tU_{i,\textcolor{ForestGreen}{l'},:}) 
\\
&+ \frac{1}{M}\sum_{
i \in [M]}\log{\sum_{\textcolor{BrickRed}
    {(j,m) \in \cN(i,l)}}  K(\tU_{i,l,:}, \tU_{\textcolor{BrickRed}{j,m},:})}\\
    \end{split}
\end{equation}

\subsection{Alignment and Uniformity coupling}
\label{ssec:coupling}
As is well known from \cite{wang2020understanding}, the InfoNCE objective optimises for two primary goals: perfect alignment between positive pairs (views of the same data point in our context) and uniformly distributing (uniformity) all the representations on the unit sphere, i.e. the first and the second term of \Cref{eq:align_unif} respectively. Although popular InfoNCE variants share the same minimisers, certain cases exhibit coupling effects between alignment and uniformity that hinder optimisation \cite{koromilas2024bridging}.

As illustrated in \Cref{fig:coupling}, the popular in practice NT-Xent \cite{chen2020simple} objective utilises interactions between different views of the same instance in both the alignment and uniformity terms. This overlap introduces a \textit{direct coupling} effect that impairs optimisation by creating conflicting objectives, ultimately delaying convergence. Further, NT-Xent also exhibit \textit{indirect coupling} since the aim in the objective is to uniformly distribute both views of all points (\ie using \(2M\) rather than \(M\) distinct points) on the unit sphere, ignoring that half of them are positives to the other half. On the other hand, DHEL \cite{koromilas2024bridging} does not include any kind of coupling.

\subsection{Mutli-View CL: Aggregating Two-View Losses}
\label{ssec:aggregations}

In SSL frameworks, some preliminary efforts have been made to leverage multiple views of the same data point within the learning process. This is typically achieved by\textit{ aggregating pairwise loss functions }$\Lpair(\tU)$, such as those of \cref{eq:mb_loss}, across different views.

\subsubsection{Pairs of views (pwe)} In the first approach \cite{tian2020contrastive, caron2020unsupervised, grill2020bootstrap, caron2021emerging}, the mean value of pairwise losses across all view combinations is computed:
\begin{equation} \Lpwe(\tU) = \frac{2}{N(N-1)} \sum_{l,
m>l}^N \Lpair([\tU_{:,l,:}, \tU_{:,m,:}])
\label{eq:pwe_loss}
\end{equation}

\subsubsection{Average (avg)} In the second approach \cite{pototzky2022fastsiam}, for each view, a mean vector based on the remaining views is calculated, and the pairwise loss is evaluated between each view and its corresponding mean vector:
\begin{equation} \Lavg(\tU) = \frac{1}{N} \sum_{l=1}^N \Lpair\left([\tU_{:,l,:}, \frac{\sum_{m \neq l}^N \tU_{:,m,:}}{N-1}]\right)
\label{eq:avg_loss}
\end{equation}

\subsubsection{PVC} In recent work, the Poly-View Contrastive (PVC) loss \cite{shidani2024polyview} was introduced to extend CL to multiple views. 
The \textit{geometric variant} of PVC, which has shown the best empirical performance in the work of \cite{shidani2024polyview}, can be simplified with the following expression (see \Cref{sec:ad_notation} for the derivation):
\begin{equation}
\label{eq:pvc}
\begin{split}
      &\Lpvc(\tU) = \frac{1}{M (N-1)}\bigg( -\sum_{l\in[N],\textcolor{ForestGreen}{l' \in[N] \setminus l} \atop i\in[M]}
      \log K(\tU_{i,l,:}, \tU_{i,\textcolor{ForestGreen}{l'},:}) + \\
    & \sum_{l, l'\in[N]\setminus l
      \atop i\in[M]} \log\big(K(\tU_{i,l,:}, \tU_{{i,l'},:}) + \sum_{\textcolor{BrickRed}{j \in [M]\setminus i \atop {m\in[N]}}}
    K(\tU_{i,l,:}, \tU_{\textcolor{BrickRed}{j,m},:})\big)\bigg).
\end{split}
\end{equation}

\begin{figure}
    \centering
    \resizebox{0.48\textwidth}{!}{
        \includegraphics{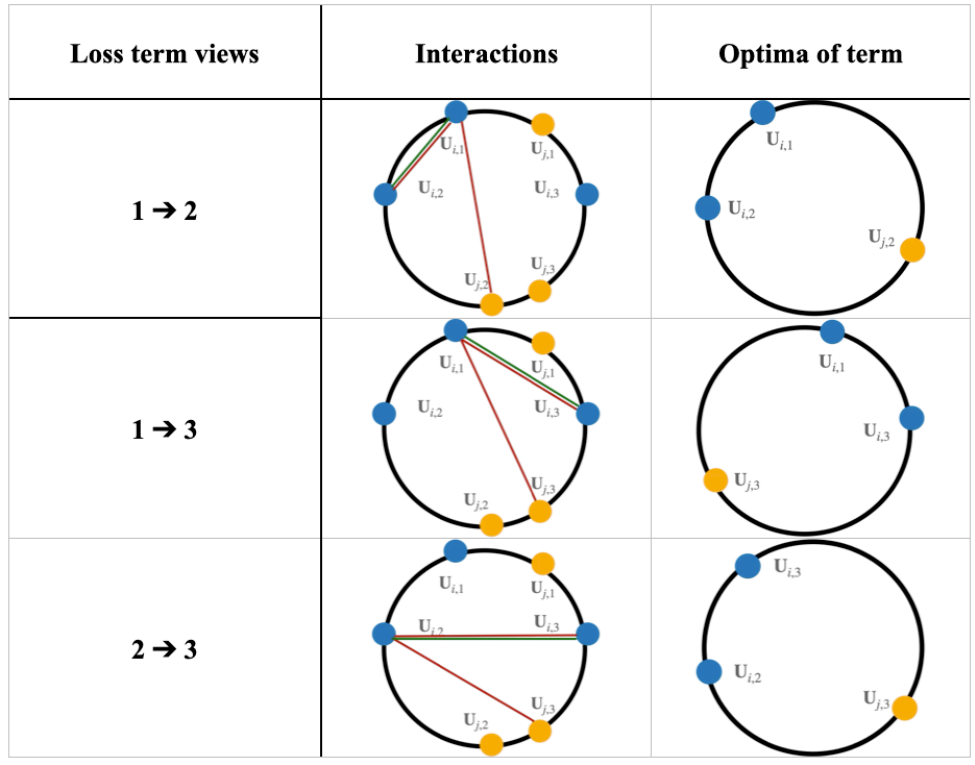}
    }
    \caption{Illustration of \textbf{conflicting optimization objectives} in pairwise loss aggregation methods (\eg PWE, Avg, PVC). Each row shows a different pairwise loss term (1→2, 1→3, 2→3) applied to \textbf{the same instance $i$}. The middle column shows how interactions are computed in each loss term between specific view pairs, where \textcolor{ForestGreen}{green lines} indicate positive interactions (alignment) and \textcolor{BrickRed}{red lines} represent negative interactions (uniformity). The right column depicts the optimal configuration for each term. Note how representation $\tU_{i,1}$ (blue) experiences conflicting forces from the 1→2 and 1→3 terms, resulting in different optimal configurations and illustrating the fundamental limitation of aggregating pairwise objectives.}
\label{fig:loss_terms}
\end{figure}

\subsection{Limitations of Aggregating Two-View Losses}
\label{ssec:limitations_of_aggregation}

Beyond the inherent limitations of pairwise loss aggregation, which \textbf{precludes direct interaction among all views} and fails to \textbf{simultaneously align all views} of the same data point, a more fundamental challenge arises in multi-view contrastive learning: \textbf{each representation must simultaneously satisfy multiple, potentially conflicting objectives}. As illustrated in \Cref{fig:loss_terms}, consider the representation $\tU_{i,1,:}$ of instance $i$ in view 1. Under pairwise aggregation, this single representation must align with both $\tU_{i,2,:}$ (from the 1$\rightarrow$2 loss term) and $\tU_{i,3,:}$ (from the 1$\rightarrow$3 loss term), while simultaneously maintaining uniformity with negative samples in view 2 and view 3 through their respective loss terms.

These objectives generate competing gradient signals: $\nabla_{\tU_{i,1,:}} \Lpair([\tU_{:,1,:}, \tU_{:,2,:}])$ and $\nabla_{\tU_{i,1,:}} \Lpair([\tU_{:,1,:}, \tU_{:,3,:}])$ may point in different directions, as each optimizes for a different pairwise relationship. The middle column of \Cref{fig:loss_terms} shows how these conflicting forces act on the same representation, while the right column depicts the ideal configuration each term seeks independently.

This conflict scales poorly with the number of views. For $N$ views, each representation participates in $(N-1)$ pairwise objectives for pwe and $N(N-1)$ for PVC, with each objective imposing its own alignment and uniformity constraints, increasing the number of conflicting interactions with the number of views. This multi-view setting creates an \textbf{overdetermined system} where satisfying one pairwise objective may compromise others. As demonstrated empirically in \Cref{sec:experiments}, this mathematical limitation manifests as degraded performance, revealing the fundamental inadequacy of pairwise aggregation for multi-view contrastive learning.

\begin{table*}[t]
\centering
\caption{Comparison of multi-view contrastive objectives for $M$ instances and $N$ views based on three principles: \textcolor{ForestGreen}{(P1)} simultaneous alignment of all views, \textcolor{BrickRed}{(P2)} accurate energy term with complete pairwise interactions, and \textcolor{NavyBlue}{(P3)} one term per instance. Only MV-InfoNCE and MV-DHEL satisfy all principles. MV-DHEL has the smaller complexity while its the only one that decouples the optimisation of alignment and uniformity.}
\label{tab:transposedcomparison}
\resizebox{\textwidth}{!}{%
\begin{tabular}{l|l|c|c|c|c|c}
\toprule
\textbf{Method} & \textbf{Objective} & \textbf{Complexity} & \makecell{\textcolor{ForestGreen}{\textbf{P1: Simultaneous}} \\ \textcolor{ForestGreen}{\textbf{Alignment}}} & \makecell{\textcolor{BrickRed}{\textbf{P2: Accurate}} \\ \textcolor{BrickRed}{\textbf{Energy Term}}} & \makecell{\textcolor{NavyBlue}{\textbf{P3: Loss Terms}} \\ \textcolor{NavyBlue}{\textbf{per Instance}}} & \makecell{\textbf{Decoupled} \\ \textbf{Alignment-Uniformity}} \\
\midrule
\textbf{pwe} & 
$\displaystyle\frac{2}{N(N-1)M} \sum_{\substack{\textcolor{NavyBlue}{l \in [N]} \\ m \in [N], m > \textcolor{NavyBlue}{l} \\ i \in [M]}} \log \left(\frac{e^{\tU_{i,\textcolor{NavyBlue}{l}}^{\top} \tU_{i,\textcolor{ForestGreen}{m}}/\tau}}{\sum_{\textcolor{BrickRed}{j \in [M]}} e^{\tU_{i,\textcolor{NavyBlue}{l}}^{\top} \tU_{\textcolor{BrickRed}{j,m}}/\tau}}\right)$ & 
$\mathcal{O}(M^2N^2)$ & 
\textcolor{BrickRed}{\ding{55}} & 
\textcolor{BrickRed}{\ding{55}} & 
\textcolor{BrickRed}{$\frac{1}{2}N(N-1)$} & 
\textcolor{BrickRed}{\ding{55}} \\
\midrule
\textbf{PVC} & 
$\displaystyle\frac{-1}{(N-1)M} \sum_{\substack{\textcolor{NavyBlue}{l\in[N]} \\ \textcolor{ForestGreen}{l' \in[N] \setminus \textcolor{NavyBlue}{l}} \\ i\in[M]}} \log \left(\frac{e^{\tU_{i,\textcolor{NavyBlue}{l}}^{\top} \tU_{i,\textcolor{ForestGreen}{l'}}/\tau}}{e^{\tU_{i,\textcolor{NavyBlue}{l}}^{\top} \tU_{i,\textcolor{ForestGreen}{l'}}/\tau} + \sum_{\substack{\textcolor{BrickRed}{m\in[N]} \\ \textcolor{BrickRed}{j \in [M]\setminus i}}} e^{\tU_{i,\textcolor{NavyBlue}{l}}^{\top} \tU_{\textcolor{BrickRed}{j,m}}/\tau}}\right)$ &
$\mathcal{O}(M^2N^3)$ & 
\textcolor{BrickRed}{\ding{55}} & 
\textcolor{BrickRed}{\ding{55}} & 
\textcolor{BrickRed}{$N(N-1)$} & 
\textcolor{BrickRed}{\ding{55}} \\
\midrule
\textbf{MV-InfoNCE} & 
$\displaystyle\frac{1}{M} \sum_{i=1}^M \log \left(\frac{\sum_{\substack{\textcolor{NavyBlue}{l\in[N]} \\ \textcolor{ForestGreen}{l' \in [N] \setminus \textcolor{NavyBlue}{l}}}} e^{\tU_{i,\textcolor{NavyBlue}{l}}^{\top} \tU_{i,\textcolor{ForestGreen}{l'}}/\tau}}{\sum_{\substack{\textcolor{NavyBlue}{l \in [N]} \\ \textcolor{BrickRed}{m \in [N]\setminus \textcolor{NavyBlue}{l}} \\ \textcolor{BrickRed}{j \in [M]}}} e^{\tU_{i,\textcolor{NavyBlue}{l}}^{\top} \tU_{\textcolor{BrickRed}{j,m}}/\tau}}\right)$ &
$\mathcal{O}(M^2N^2)$ & 
\textcolor{ForestGreen}{\ding{51}} & 
\makecell{\textcolor{ForestGreen}{\ding{51}}} & 
\textcolor{ForestGreen}{\textbf{1}} & 
\textcolor{BrickRed}{\ding{55}} \\
\midrule
\textbf{MV-DHEL} & 
$\displaystyle\frac{1}{M} \sum_{i=1}^M \log \left(\frac{\sum_{\substack{\textcolor{NavyBlue}{l\in[N]} \\ \textcolor{ForestGreen}{l' \in [N] \setminus \textcolor{NavyBlue}{l}}}} e^{\tU_{i,\textcolor{NavyBlue}{l}}^{\top} \tU_{i,\textcolor{ForestGreen}{l'}}/\tau}}{\prod_{\textcolor{NavyBlue}{l \in [N]}} \sum_{\textcolor{BrickRed}{j \in [M]}} e^{\tU_{i,\textcolor{NavyBlue}{l}}^{\top} \tU_{\textcolor{BrickRed}{j,\textcolor{NavyBlue}{l}}}/\tau}}\right)$ &
$\mathcal{O}(M^2N)$ & 
\textcolor{ForestGreen}{\ding{51}} & 
\makecell{\textcolor{ForestGreen}{\ding{51}}} & 
\textcolor{ForestGreen}{\textbf{1}} & 
\textcolor{ForestGreen}{\ding{51}} \\
\bottomrule
\end{tabular}
}
\end{table*}

 \section{Multi-View Contrastive Objectives}\label{sec:method}

\subsection{A Framework for Multi-View Contrastive Losses}
We first establish a framework that generalizes each term of \Cref{eq:align_unif} to multiple views by incorporating $N$-wise interactions. 
\subsubsection{Framework Principles}
To properly extend contrastive losses to multiple views, we define three fundamental principles that must be satisfied for a theoretically sound multi-view contrastive objective:

\noindent\textcolor{ForestGreen}{\textbf{P1: Simultaneous Alignment}}. Given a data point $i$ and a view of interest $l$, all other views of the same data point must be simultaneously aligned within a single term of the objective. This ensures that representations become \textit{invariant to all transformations} simultaneously, avoiding competing optimization terms that could lead to suboptimal solutions or training instability.

\noindent\textcolor{BrickRed}{\textbf{P2: Accurate Energy Term}}. Contrastive objectives fundamentally optimize for alignment and uniformity \cite{wang2020understanding}. The uniformity term corresponds to minimizing the energy of a point configuration $\{\mathbf{u}_1, \ldots, \mathbf{u}_M\}$, a set of representations in our case. Specifically, we seek to minimize the total pairwise energy $\sum_i \sum_j K(\mathbf{u}_i, \mathbf{u}_j)$ \cite{koromilas2024bridging}. As shown in \cite{wang2020understanding}, this energy minimization is equivalent to minimizing $\sum_i \log \sum_j K(\mathbf{u}_i, \mathbf{u}_j)$, which forms the uniformity term. To maintain theoretical consistency with contrastive objectives, the negative set used in uniformity calculations must represent a complete point configuration that captures \textit{all possible pairwise interactions}. For instance, the PVC loss in \Cref{eq:pvc} violates this principle by omitting certain interactions (\eg $\tU_{i, l}^T\tU_{i,m}, m \neq l'$ is not calculated inside the log), resulting in an incomplete energy term.

\noindent\textcolor{NavyBlue}{\textbf{P3: One Term per Data Instance}}. As discussed in \Cref{ssec:limitations_of_aggregation}, current multi-view contrastive losses utilize multiple optimization terms per instance, introducing competing objectives for the same data point. A principled extension to multiple views should maintain one term per instance for the complete objective (\ie, alignment + uniformity), consistent with two-view objectives, to ensure stable and efficient optimization.

Our framework defines multi-view loss functions that satisfy all three principles through three key design components: (i) the placement of the \textcolor{NavyBlue}{summation over the view of interest} $\textcolor{NavyBlue}{l \in [N]}$ (indicated in \textcolor{NavyBlue}{blue}) either inside or outside the logarithm controls \textcolor{NavyBlue}{P3}; (ii) the \textcolor{ForestGreen}{positive index set $\mathcal{P}(l)$}, which determines how to sample positives for view $\textcolor{NavyBlue}{l}$ from the same data instance, controls \textcolor{ForestGreen}{P1}; and (iii) the \textcolor{BrickRed}{negative set $\mathcal{N}(i,l)$}, which specifies different instances and their views for uniformity calculation, controls \textcolor{BrickRed}{P2}.

\subsubsection{Multi-view Alignment} First off, we observe that generalising the first term requires simultaneously aligning all views of each datapoint. To achieve this, we need to extend the formulation from the single different view $l'$ to a set of (typically all) different views denoted with $\textcolor{ForestGreen}{\cP(l)}$. Second we need to choose whether to apply the \textcolor{NavyBlue}{summation} inside or outside the logarithm. This leads to the following two alignment terms:

\begin{align}
L_{\text{align}}(\tU)  &=
     \frac{-1}{NM} \sum_{
     {\textcolor{NavyBlue}{l \in [N]}} \atop i \in [M]}
\log{\big(\sum_{\textcolor{ForestGreen}{l' \in \cP(l)}}^N   K(\tU_{i,
\textcolor{NavyBlue}{l},:}, \tU_{i,\textcolor{ForestGreen}{l'},:}) \big)}\label{eq:mv_align_out}
\\
L_{\text{align}}(\tU)  &=
     \frac{-1}{NM} \sum_{i \in [M]}
\log{\big(\sum_{\textcolor{ForestGreen}{\textcolor{NavyBlue}{l \in [N]} \atop l' \in \cP(l)}}^N   K(\tU_{i,
\textcolor{NavyBlue}{l},:}, \tU_{i,\textcolor{ForestGreen}{l'},:}) \big)}\label{eq:mv_align_in}
\end{align}

Observe that for a given view of interest $\textcolor{NavyBlue}{l}$, contrary to previous objectives in \Cref{ssec:aggregations}, the summation of positive views occurs \textit{inside} the logarithm, promoting the desired simultaneous alignment to the view of interest. However, \Cref{eq:mv_align_out} instroduces multiple loss terms per data instance, which as discussed in \Cref{ssec:limitations_of_aggregation} hurts optimization by introducing competing objectives for the same data point.

\subsubsection{Multi-view Uniformity}Generalizing the uniformity term is straightforward as it requires contrasting each single-view datapoint to a set of negatives. Our definition of the negative set $\textcolor{BrickRed}{\cN(i,l)}$ is sufficient in this case. Therefore, by placing the view of interest $\textcolor{NavyBlue}{l}$ either outside or inside the logarithm we obtain:
\begin{align}
L_{\text{unif}}(\tU)  &=
     \frac{1}{NM} \sum_{
     {\textcolor{NavyBlue}{l \in [N]}} \atop i \in [M]}
\log{\big(\sum_{\textcolor{BrickRed}{(j,m) \in \cN(i,l)}} K(\tU_{i,\textcolor{NavyBlue}{l},:}, \tU_{\textcolor{BrickRed}{j,m},:}) \big)} \label{eq:mv_unif_out} \\
L_{\text{unif}}(\tU)  &=
     \frac{1}{NM} \sum_{i \in [M]}{ \log{\big(\sum_{\textcolor{NavyBlue}{l \in [N]} \atop \textcolor{BrickRed}{(j,m) \in \cN(i,l)}}} K(\tU_{i,\textcolor{NavyBlue}{l},:}, \tU_{\textcolor{BrickRed}{j,m},:}) \big)} \label{eq:mv_unif_in}
\end{align}

These equations differ from past objectives since each single-view datapoint is contrasted against \textit{all} its counterparts (unlike pairwise aggregations). Again, here \Cref{eq:mv_unif_out}, if optimized independently, introduces $N$ different optimization terms for each data point while \Cref{eq:mv_unif_in} calculates one term per data instance.

\subsection{Multi-View Extensions}
Now all contrastive losses can be expressed by setting, the \textcolor{NavyBlue}{place of summation} of the view of interest, the positive $\textcolor{ForestGreen}{\cP(l)}$ and the negative $\textcolor{BrickRed}{\cN(i,l)}$ sets. 

\subsubsection{Multi-view InfoNCE} By \textcolor{NavyBlue}{summing inside the logarithm} in order to utilize one optimization term per data point and also setting $\textcolor{ForestGreen}{\cP(l) = l' \in [N] \setminus l}$ and $\textcolor{BrickRed}{\cN(i,l) = \{(j,m) \mid j \in [M], m \in [N] \setminus l\}}$ as the index sets, we obtain \textit{MV-InfoNCE}, an extension that is based on the classical InfoNCE. 

\begin{equation}
    \begin{split}
    L_{\text{MV-InfoNCE}}(\tU) &= 
\frac{1}{M} \sum_{i \in [M]} -\log{\sum_{\textcolor{NavyBlue}{l\in[N]} \atop
 \textcolor{ForestGreen}{l' \in [N] \setminus l}} K(\tU_{i,\textcolor{NavyBlue}{l},:}^{\top} \tU_{i, \textcolor{ForestGreen}{l'},:})}\\
& +\frac{1}{M} \sum_{i \in [M]}\log{\sum_{\textcolor{NavyBlue}{l\in[N]} \atop {\textcolor{BrickRed}{j \in [M] \atop m \in [N]\setminus l}}}K(\tU_{i,\textcolor{NavyBlue}{l},:}^{\top} \tU_{\textcolor{BrickRed}{j,m},:})}
\end{split}
\label{eq:mvinfonce_mb}
\end{equation}

Observe that we simultaneously consider all single-view datapoints interactions in the computations of both the alignment (\textcolor{ForestGreen}{P1}) and uniformity (\textcolor{BrickRed}{P2}) terms, while also computing one term per data point (\textcolor{NavyBlue}{P3}). This objective naturally extends InfoNCE family to multiple views. Following the NT-Xent formulation, we generalize its two-view objective which (i) uses one term per datapoint, (ii) simultaneously aligns both views of the instance, (iii) calculates all interactions between all views of the data point of interest and all views of every other data point in the negative index set.

However, this objective inherits the drawbacks of its parent (InfoNCE), were there is a coupling in the optimization of alignment and uniformity (see \cref{ssec:coupling}). When scaling to more views, the aim is to uniformly distribute all datapoints views, but the objective disregards that some are different views of the \textit{same} data point. That is, \textbf{using more views introduces further coupling} thus hurting the optimisation of the joint objective.

\subsubsection{Multi-view DHEL} To address the growth in coupling, we propose extending DHEL, an objective that naturally decouples alignment from uniformity, to the multiview setup. By using the following index sets $\textcolor{ForestGreen}{\cP(l) = l' \in [N] \setminus l}$ and $\textcolor{BrickRed}{\cN(i,l) = \{(j,m) \mid j \in [M]\setminus i, m = l\}}$ we end up with \textit{negative interactions only in the same view} which is sufficient in order to avoid coupling for two-view set-ups. However, when utilizing more than two views choosing the \textcolor{NavyBlue}{place of summation} of the view of interest is crucial.

By avoiding interactions that cause coupling, \ie by plugging $\textcolor{BrickRed}{\cN(i,l) = \{(j,m) \mid j \in [M]\setminus i, m = l\}}$ into \Cref{eq:mv_unif_in} we end up with an objective that does not calculate all interactions among all summads and thus it violates \textcolor{BrickRed}{P2} since it does not calculate an energy term. Contrary, summing outside the log introduces N different energy/uniformity terms, each calculating the energy of all data instances under a specific view. We further note that summing outside the log in the uniformity term combined with summing inside the log in the alignment term creates an objective that has the desired behavior of utilizing one term per data instance. Thus by using \Cref{eq:mv_align_in} and \Cref{eq:mv_unif_out} along our positive and negative index sets, we get the desired \textit{MV-DHEL}:

\begin{equation}
\begin{split}
    L_{\text{MV-DHEL}}(\tU)
&= \frac{1}{M} \sum_{i \in [M]} -\log{\sum_{\textcolor{NavyBlue}{l\in[N]} \atop
 \textcolor{ForestGreen}{l' \in [N] \setminus l}} K(\tU_{i,\textcolor{NavyBlue}{l},:}^{\top} \tU_{i,\textcolor{ForestGreen}{l'},:})}
\\&
+\frac{1}{M} \sum_{\textcolor{NavyBlue}{l \in [N]} \atop i \in [M]}\log{\sum_{\textcolor{BrickRed}{j \in [M]\setminus i}}K(\tU_{i,\textcolor{NavyBlue}{l},:}^{\top} \tU_{\textcolor{BrickRed}{j,l},:})} 
\end{split}
\label{eq:mvdhel_mb}
\end{equation}

This objective calculates the alignment across all views of a data point as before. However, \textit{ the uniformity term considers each view of a datapoint separately}. In this way MV-DHEL optimises only for in-view uniformity without compromising alignment and thus \textbf{effectively decoupling the optimisation of alignment and uniformity}. See \Cref{fig:coupling} for an illustration.

\subsection{Mutli-View Losses Comparison}

\Cref{tab:transposedcomparison} provides a comprehensive comparison of multi-view contrastive objectives against our three fundamental principles. The comparison reveals critical limitations in existing approaches and validates our proposed extensions.

\textbf{Existing methods fail to satisfy the principles.} Pairwise aggregation methods (pwe) and PVC violate all three principles. For \textcolor{ForestGreen}{P1}, both methods process views pairwise, computing one term per positive pair rather than simultaneously aligning all views. For \textcolor{BrickRed}{P2}, while both methods include all views of the same data point in their point configuration, they fail to capture the complete energy by excluding critical pairwise interactions from the energy term—specifically, interactions like $\mathbf{\tU}_{i,l}^T\mathbf{\tU}_{i,m}$ for $m \neq l'$ are omitted from the denominator, resulting in an incomplete energy minimization. For \textcolor{NavyBlue}{P3}, both methods generate multiple optimization terms per instance: pwe produces $\frac{1}{2}N(N-1)$ terms and PVC produces $N(N-1)$ terms, introducing competing objectives that destabilize optimization for each data point.

\textbf{MV-InfoNCE and MV-DHEL satisfy all principles.} Our framework yields two theoretically sound objectives. MV-InfoNCE extends the InfoNCE family by placing the summation over views inside the logarithm, achieving simultaneous alignment of all views (\textcolor{ForestGreen}{P1}), complete pairwise interactions (\textcolor{BrickRed}{P2}), and one term per instance (\textcolor{NavyBlue}{P3}). MV-DHEL employs a hybrid approach—summing inside the logarithm for alignment but outside for uniformity—which, when combined with calculating view-specific uniformity not only satisfies all principles but also decouples the optimization of alignment and uniformity.

\textbf{MV-DHEL vs Mv-InfoNCE.} While both MV-InfoNCE and MV-DHEL satisfy all principles, MV-DHEL needs less computations and completely decouples the optimization of alginment and uniformity. This efficiency and optimization gain becomes increasingly important as the number of views $N$ grows, making MV-DHEL particularly attractive for applications with many augmentations or modalities.

\textbf{Other Losses.} In \Cref{tab:mv_methods_comparison} and \Cref{tab:results_mm_supp} in the appendix, we further evaluate two different losses which are obtained by using the alignment to \cref{eq:mv_align_out}, the uniformity to \cref{eq:mv_unif_out} and the negative index sets based on the corresponding sets of MV-InfoNCE and MV-DHEL respectively.

\subsection{Asymptotic Optima of Multi-View Losses}
In the following section, we theoretically analyse the minima of our \textit{expected} multi-view objectives and show that, asymptotically (as the number of negatives tends to infinity), they are optimised by perfect alignment (all views share the same representation) and perfect uniformity (representations obey a uniform law on the unit sphere).

We will denote the pushforward measures induced by $f$ (encoder) with $f_\#\pdata$. Additionally, we denote with $p_{\text{trans}}$ the distribution of a datapoint sampled by $p_{\text{init}}$ and then transformed by a single transformation sampled by $p_T$.
By sampling a tensor of $M$ datapoints of \( N \) positive views from the pushforward measure induced by \( f \) on the $N$-view distribution, \ie \( \bU_{j} = (\bu_1, \dots, \bu_N) \overset{\text{i.i.d}}{\sim} \funsimple_\#\pdata \), $j \in [M]$ we establish the expectations of \cref{eq:mvinfonce_mb} and \cref{eq:mvdhel_mb} for the gaussian kernel:
\begin{equation}
\begin{split}
&E_1 = 
\underset{
\bU_{j}\overset{\text{i.i.d}}{\sim} \funsimple_\#\pdata^{M}}{\E}\left[-\log \left(\frac{\sum_{l \in [N], \atop l' \in [N]\setminus l} e^{\bU_{1,l}^{\top} \bU_{1,l'}{/\tau}}}{\sum_{l \in [N] \atop {j\in [M]\atop m \in [N]\setminus l}} e^{\bU_{1,l}^{\top} \bU_{j, m}{/\tau}}}\right)\right]
\end{split}
\label{eq:mvinfonce_expectation}
\end{equation}

\begin{equation}
\begin{split}
&E_2 =  \underset{
\bU_{j}\overset{\text{i.i.d}}{\sim} \funsimple_\#\pdata^{M}}{\E}\left[-\log \left(\frac{\sum_{l \in [N], \atop l' \in [N]\setminus l}  e^{\bU_{1,l}^{\top} \bU_{1,l'}{/\tau}}}{
\prod_{l \in [N]}\sum_{{j \in  [M-1]}} e^{\bU_{1,l}^{\top} \bU_{j, l}{/\tau}}}\right)\right]
\end{split}
\label{eq:mvdhel_expectation}
\end{equation}

\begin{theorem}
The expectations of the following batch-level contrastive loss functions: $L_{\text{MV-InfoNCE}}(\cdot)$, $\Lmvdhel(\cdot)$
have the \textbf{same asymptotic behaviour} when subtracting appropriate normalising constants 
 \ie\ when $M \to \infty$ they converge to the asymptotic formula of InfoNCE \cite{wang2020understanding}:
\begin{equation}\label{eq:asymptotic}
\underset{(\bu, \bv)\sim f_\#p_{\text{trans}}^2}{\E}
\left[-\bv^\top \bu/\tau\right] + \underset{\bv{\sim} f_\#p_{\text{trans}}}{\E}\left[\log \underset{ \bu{\sim} f_\#p_{\text{trans}}}{\E}\left[e^{\bv^\top \bu/\tau}\right]\right].
\end{equation}
\end{theorem}
This implies that, \textit{if there exists an encoder that achieves perfect alignment and uniformity then it forms the only minimiser of the objectives in \Cref{eq:mvinfonce_expectation} and \Cref{eq:mvdhel_expectation}.}
\begin{table*}[ht!]
\caption{Linear probing based performance comparison with accuracy and improvement (Diff) based on previous the using one lesser view for each method. Green values indicate the best performance per dataset, bold values indicate the highest values per view.}
\label{tab:results}
\centering
\resizebox{\textwidth}{!}{ 
\renewcommand{\arraystretch}{1.2} 
\begin{tabular}{|c|c|cc|cc|cc||cc|cc|}
\hline
\textbf{Dataset} & \textbf{\# Views} & \multicolumn{2}{c|}{\textbf{pwe} (Eq. \ref{eq:pwe_loss})} & \multicolumn{2}{c|}{\textbf{avg} (Eq. \ref{eq:avg_loss})} & \multicolumn{2}{c||}{\textbf{PVC} (Eq. \ref{eq:pvc})} & \multicolumn{2}{c|}{\textbf{MV-InfoNCE} (Eq. \ref{eq:mvinfonce_mb})} & \multicolumn{2}{c|}{\textbf{MV-DHEL} (Eq. \ref{eq:mvdhel_mb})} \\ \hline
\textbf{} & \textbf{} & \textbf{Accuracy} & \textbf{Diff} & \textbf{Accuracy} & \textbf{Diff} & \textbf{Accuracy} & \textbf{Diff} & \textbf{Accuracy} & \textbf{Diff} & \textbf{Accuracy} & \textbf{Diff} \\ \hline
\rowcolor[HTML]{EFEFEF} 
\multirow{3}{*}{\textbf{CIFAR10}}      & 2        &   86  & -- &   86  & -- &     85.8      & --       &       86    & --       &     \textbf{87.4} & --        \\
                                       & 3        &   87.5  & +1.5   &  87.4   & +1.4  &      87.0    & +1.2    &     87.8      & \textbf{+1.8}       &    \textbf{89.1}   & +1.7        \\
                                       & 4        &   88.7  & +1.2   &  88.2   & +0.8  &      88.0    & \textbf{+1.0}    &     88.8    & \textbf{+1.0}       &      {\color{ForestGreen}\textbf{89.5}}  & +0.4         \\ \hline
\rowcolor[HTML]{EFEFEF} 
\multirow{3}{*}{\textbf{CIFAR100}}     & 2        & 58.1    & -- &   58.1 & --  & 57.3 & --   &    58.1  & --         &    \textbf{59.4}   & --       \\
                                       & 3        &  59.9  & +1.8   &   60.4 & +2.3   & 60.3 & +3.0 &      60.6 & \textbf{+2.5}     &    \textbf{61.8}   & +2.4       \\
                                       & 4        &   60.9  & +1.0   &   60.8  & +0.4  & 61.1 & +0.8 &      61.2 & +0.6    & {\color{ForestGreen}\textbf{62.7}} & \textbf{+0.9} \\ \hline
\rowcolor[HTML]{EFEFEF} 
\multirow{3}{*}{\textbf{ImageNet-100}} & 2        &  72.2   & -- &  72.2   & -- &      72.2     & --        &     72.2      & --       &     \textbf{73.3}  & --       \\
                                       & 3        &   75  &   +2.8  &   74.8  &   +2.6  &       75    &    +2.8       &     75.2     &     +3        &      \textbf{77.1}    &    \textbf{+3.8}      \\
                                       & 4        &   73.9   &  {\color{BrickRed}-1.1}   &   73.7  &   {\color{BrickRed}-1.1}  &     74.4      &     {\color{BrickRed}-0.6}      &      75.8     &      \textbf{+0.6}      &    {\color{ForestGreen}\textbf{77.2}}   &     +0.1    \\ \hline
\rowcolor[HTML]{EFEFEF} 
\multirow{3}{*}{\textbf{ImageNet-1K}}  & 2        &  60   & -- &  60   & -- &     59.7      & --        &      60     & --         &      \textbf{61.2}     & --         \\
                                       & 3        &  61.2   &  +1.2   &  61   &  +1   &        61.4   &       +\textbf{1.7}    &     60.8      & +0.8           & \textbf{61.9}          & + 0.7          \\
                                       & 4        &    62 &    + 0.8 &    61.6 &  +0.6  &  62.4         &   +\textbf{0.7}        &   61.2        &    + 0.4       &      {\color{ForestGreen}\textbf{62.6}}     & + \textbf{0.7}          \\ \hline
\end{tabular}
}
\end{table*}
\section{Experimental Evaluation}
\label{sec:experiments}

We empirically validate our proposed objectives (Eq. \ref{eq:mvinfonce_mb} and \ref{eq:mvdhel_mb}) by benchmarking against three established techniques as described in \cref{ssec:aggregations}. These methods are as follows: (i) \textbf{pwe} (Eq. \ref{eq:pwe_loss}) — aggregation based on the pairwise loss computed for all pairs of views; (ii) \textbf{avg} (Eq. \ref{eq:avg_loss}) — pairwise loss between each view and the mean vector of the remaining views; and (iii) \textbf{PVC} (Eq.  \ref{eq:pvc}) as proposed in \cite{shidani2024polyview}.

\subsection{Downstream Performance}
Experiments are conducted on four standard image classification datasets: \textit{CIFAR10, CIFAR100, ImageNet-100, and ImageNet1K}, following common SSL practices \cite{wang2021unsupervised, yeh2022decoupled, zhang2022dual, wang2020understanding, chen2020simple}. We use ResNet50 for ImageNet-100/ImageNet1K and ResNet18 for CIFAR10/CIFAR100. Models are trained for 100 epochs on ImageNet1K (batch size 512, SGD optimizer) and 200 epochs on other datasets (batch size 256, SGD). Further implementation details are provided in \cref{sec:impl_details} in the appendix.

\subsubsection{Linear Separability} As is common in the SSL literature \cite{wang2021unsupervised, yeh2022decoupled, zhang2022dual, wang2020understanding, chen2020simple} we evaluate the linear separability of the representation space by training a linear classifier on freezed representations (linear probing) for 100 epochs \cite{wang2021unsupervised}.

In \Cref{tab:results}, we compare the performance of all five methods across four datasets with varying view multiplicities (2, 3, and 4). Our results demonstrate the clear advantages of our proposed approaches. \textbf{MV-DHEL} consistently achieves the \textbf{highest overall accuracy} across all datasets, while \textbf{MV-InfoNCE} exhibits the greatest \textbf{performance scaling} as views increase. Unlike baseline methods, both our approaches show significant accuracy gains when advancing from two to four views, confirming their effectiveness in leveraging multi-view information.

\begin{table}[ht!]
\caption{Top-1 accuracy for weighted k-nearest neighbor classification (k=200)}
\label{tab:knn}
\centering
\resizebox{0.48\textwidth}{!}{
\begin{tabular}{|c|c|c|c|c|c|}
\hline
\textbf{Dataset} & \textbf{\# Views} & {\textbf{pwe}} & {\textbf{PVC}} & {\textbf{MV-InfoNCE}} & {\textbf{MV-DHEL}} \\ \hline
\multirow{2}{*}{\textbf{CIFAR10}} 
& 3 & 82.3 & 82.3 & 83 & \textbf{\textcolor{ForestGreen}{85.1}} \\
& 4 & 83.4 & 83.6 & 84.3 & \textbf{\textcolor{ForestGreen}{86.7}} \\ \hline
\multirow{2}{*}{\textbf{CIFAR100}} 
& 3 & 43.9 & 43.8 & 44.9 & \textbf{\textcolor{ForestGreen}{49.9}} \\
& 4 & 45.1 & 45.3 & 45.9 & \textbf{\textcolor{ForestGreen}{51.8}} \\ \hline
\multirow{2}{*}{\textbf{ImageNet-100}} 
& 3 & 65.3 & 63.8 & 64.7 & \textbf{\textcolor{ForestGreen}{68.5}} \\
& 4 & 63.3 & 65.6 & 65.9 & \textbf{\textcolor{ForestGreen}{70.1}} \\ \hline
\multirow{2}{*}{\makecell{\textbf{ImageNet-1K}}} 
& 3 & 42.1   & 42.4   & 42.5   & \textbf{\textcolor{ForestGreen}{42.6}} \\
& 4 & 43.8   & 43.7 & 44.9 & \textbf{\textcolor{ForestGreen}{46.3}} \\ \hline
\end{tabular}}
\end{table}

\subsubsection{Neighborhood-Based Separability}

Here, we evaluate the learned representations using weighted k-nearest neighbor classification (k=200) following \cite{wu2018unsupervised}, which reflects the local neighborhood structure in the feature space. \Cref{tab:knn} shows that both our proposed methods significantly outperform existing baselines, with MV-DHEL achieving the highest accuracy across all datasets and view configurations. The performance gains are substantial: on ImageNet-100 with 4 views, MV-DHEL reaches 70.1\% accuracy compared to 65.6\% for PVC. These results demonstrate that our methods learn representations with superior neighborhood separability.

\subsection{Application to Multimodal Data}

\begin{table*}[ht!]
\caption{Performance comparison on multimodal sentiment analysis datasets. The supervised method provides an upper bound reference. Values in parentheses indicate absolute improvement over the best baseline.}
\label{tab:results_multimodal}
\centering
\begin{tabular}{|l|l|c|c|c|c||c|c|}
\hline
\multirow{2}{*}{\textbf{Dataset}} & \multirow{2}{*}{\textbf{Metric}} & \multicolumn{6}{c|}{\textbf{Methods}} \\
\cline{3-8}
 &  & \textbf{Supervised} & \textbf{PWE} & \textbf{AVG} & \textbf{PVC} & \textbf{MV-InfoNCE} & \textbf{MV-DHEL} \\
\hline
\multirow{2}{*}{\textbf{CMU-MOSEI}} 
& \textbf{Accuracy}($\uparrow$) & \cellcolor[HTML]{EFEFEF}82.9 & 75.4 & 75.7 & 74.3 & 76.8 (+1.1) & {\color{ForestGreen}\textbf{79.6} (+3.9)} \\
& \textbf{MAE}($\downarrow$) & \cellcolor[HTML]{EFEFEF}0.587 & 0.707 & 0.713 & 0.755 & 0.708 (-0.001) & {\color{ForestGreen}\textbf{0.668} (-0.039)} \\
\hline
\multirow{2}{*}{\textbf{CH-SIMS}} 
& \textbf{Accuracy}($\uparrow$) & \cellcolor[HTML]{EFEFEF}83.2 & 76.1 & 76.3 & 68.1 & 76.6 (+0.3) & {\color{ForestGreen}\textbf{79.4} (+3.1)} \\
& \textbf{MAE}($\downarrow$) & \cellcolor[HTML]{EFEFEF}0.354 & 0.448 & 0.468 & 0.563 & 0.421 (-0.027) & {\color{ForestGreen}\textbf{0.392} (-0.056)} \\
\hline
\end{tabular}
\end{table*}

Unlike views that share the same underlying distribution, modalities originate from distinct ones. By using separate encoders for each modality, their representations can be treated as alternative views of the same data point, allowing direct application of contrastive learning to multimodal data (e.g., CLIP \cite{radford2021learning}). However, CL methods are mainly designed for bimodal settings and struggle with scaling to multiple modalities \cite{ruan2024tricolo, liu2021contrastive, sun2024contextual}. Here we empirically evaluate the effectiveness of our method to multimodal setups.

We apply our methods to Multimodal Sentiment Analysis (MSA), a well-established multimodal task \cite{liang2023multizoo} that integrates three heterogeneous modalities: audio, vision, and text. MSA is typically treated as a regression task \cite{mao2022msena}, where models predict sentiment polarity on a continuous scale. Final predictions are then thresholded into discrete categories, with corresponding classification metrics reported (binary Accuracy here).

We evaluate our approach on two datasets with distinct multimodal characteristics: CMU-MOSEI \cite{bagher-zadeh-etal-2018-multimodal}, the largest multimodal sentiment benchmark comprising 65 hours of multimodal sentiment data, and CH-SIMS \cite{yu2020ch}. These datasets exhibit fundamentally different patterns of cross-modal correlation with respect to downstream tasks. In CMU-MOSEI, the textual modality dominates, showing strong correlation with task performance, while the audio and visual modalities provide only marginal improvements. In contrast, CH-SIMS demonstrates high correlation across all modalities, with strong inter-modal agreement—for instance, 86\% concordance between audio and multimodal annotations \cite{yu2020ch}. This high modality alignment in CH-SIMS yields substantial mutual information across views, a property particularly advantageous for contrastive learning approaches. Full implementation details are provided in Section \ref{ssec:impl_details}.

\Cref{tab:results_multimodal} demonstrates that, again, \textbf{MV-DHEL} \textbf{performs significantly higher} in the multimodal setup, greatly outperforming other methods. The simultaneous utilization of all multimodal interactions places MV-InfoNCE second, while pairwise loss aggregations (pwe and avg) yield slightly lower performance. An interesting observation is that PVC performs notably worse, suggesting it is not suited for multimodal learning. This is reasonable, as PVC focuses on learning from pairs based on their mutual information. This is inefficient for multimodal settings, where the information across modalities is heterogeneous and not always consistent across data points (e.g., text may contain more information than other modalities in an instance).

\subsection{Ablation studies}

\begin{figure*}[t!]
    \centering
   \begin{minipage}{0.24\textwidth}
    \resizebox{\textwidth}{!}{\includegraphics{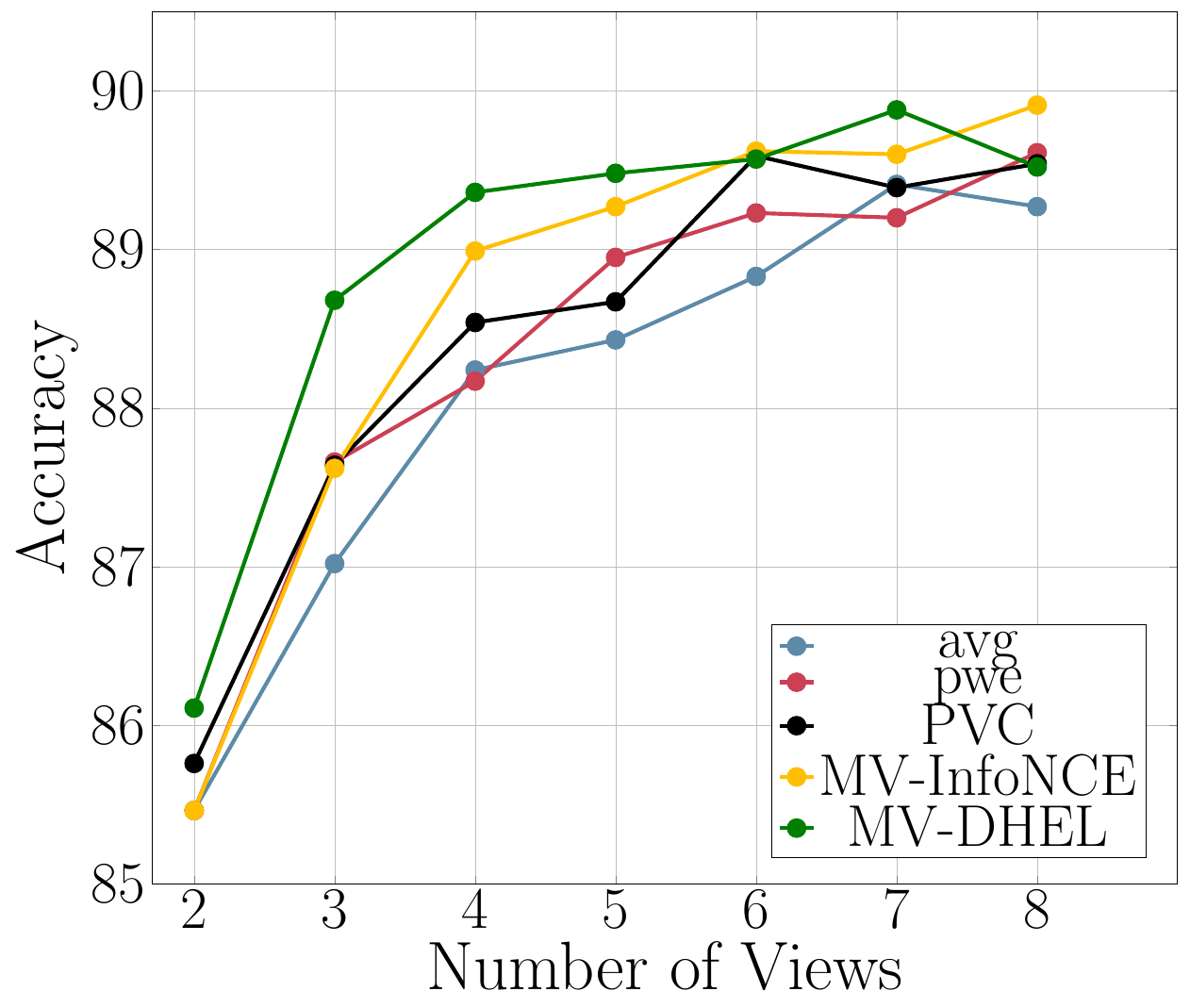}}
    \end{minipage}
    \begin{minipage}{0.24\textwidth}
    \resizebox{\textwidth}{!}{\includegraphics{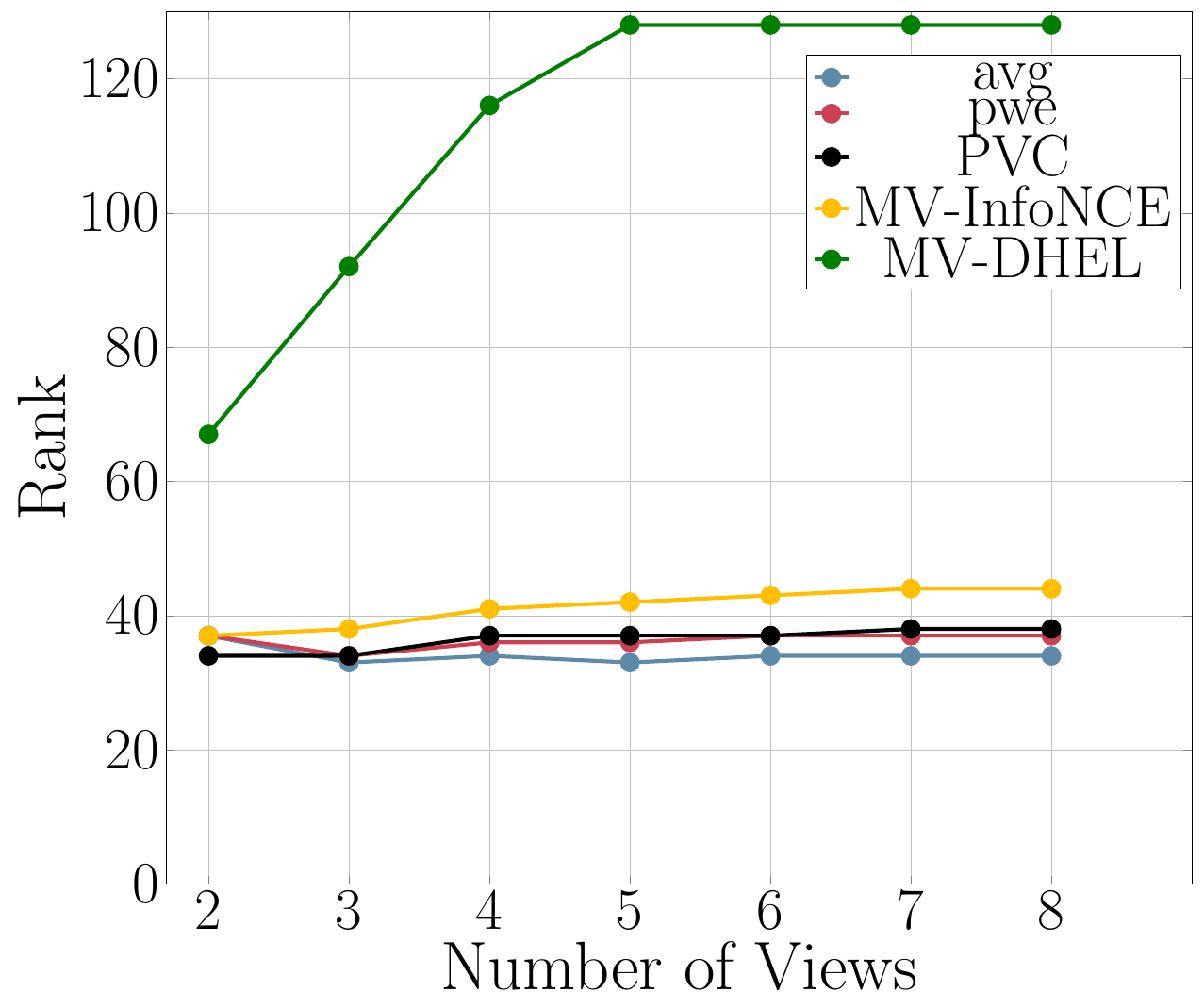}}
    \end{minipage}
    \begin{minipage}{0.24\textwidth}
    \resizebox{\textwidth}{!}{\includegraphics{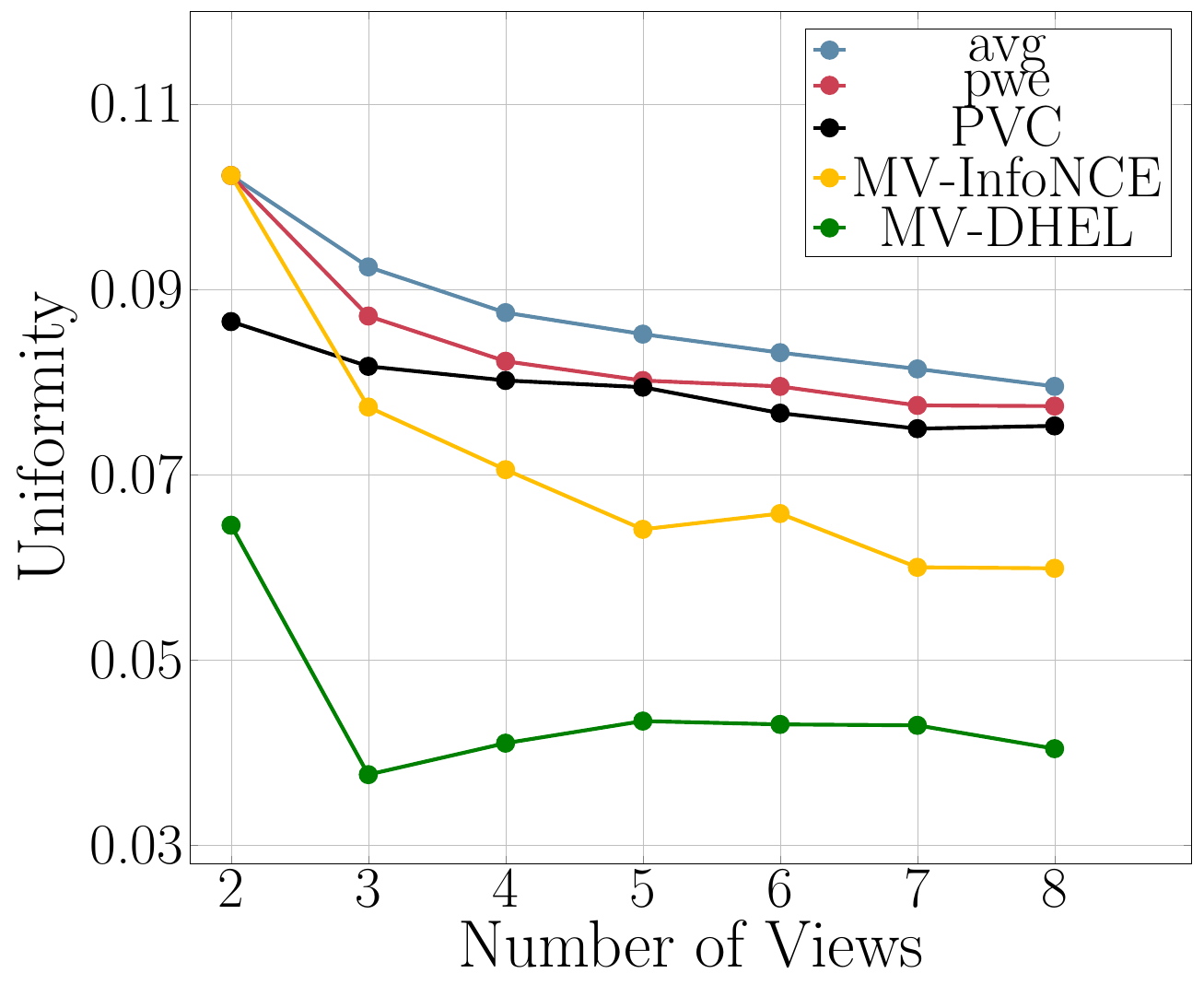}}
    \end{minipage}
    \begin{minipage}{0.23\textwidth}
    \resizebox{\textwidth}{!}{\includegraphics{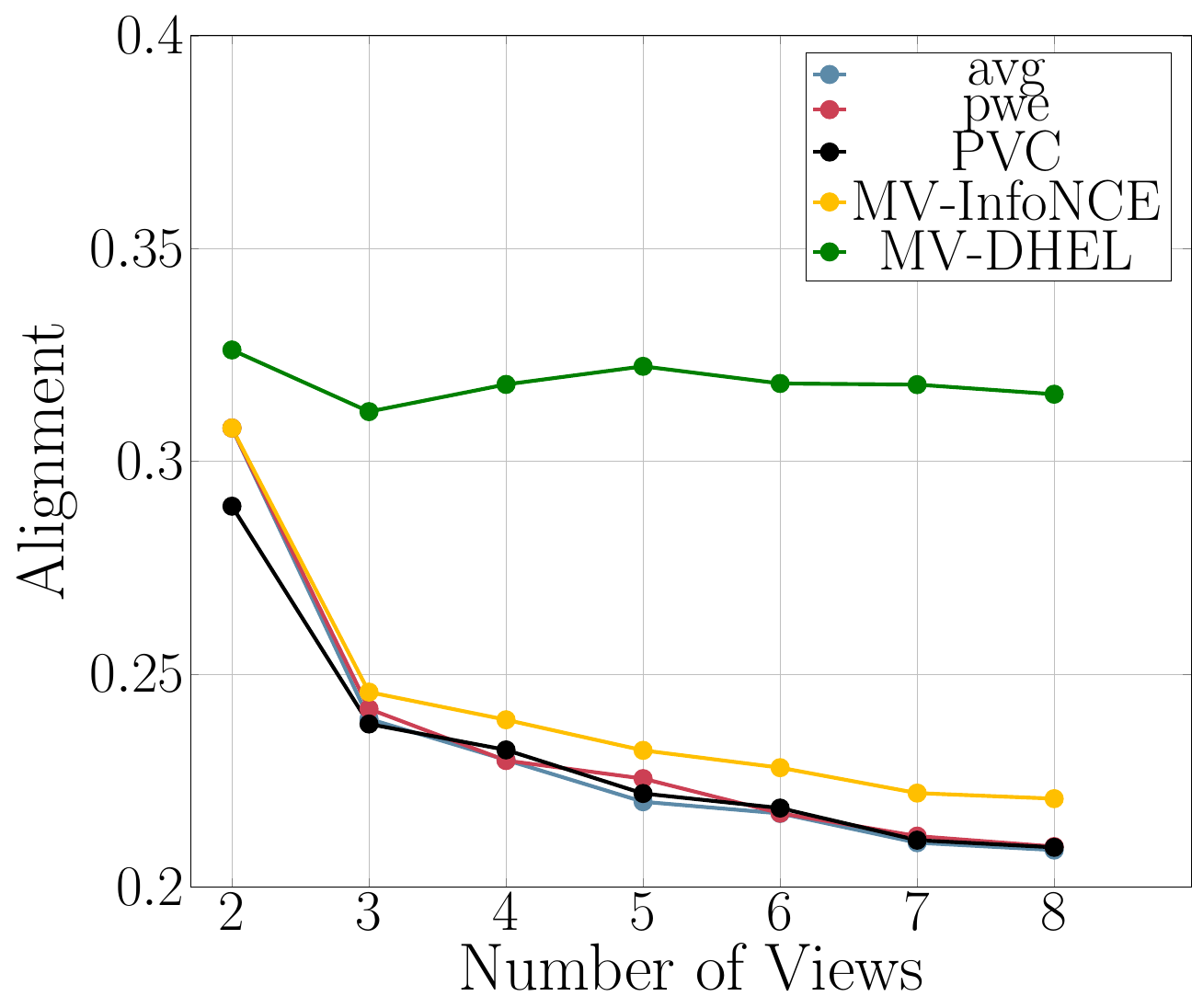}} 
    \end{minipage}
    \begin{minipage}{0.23\textwidth}
    \resizebox{\textwidth}{!}{\includegraphics{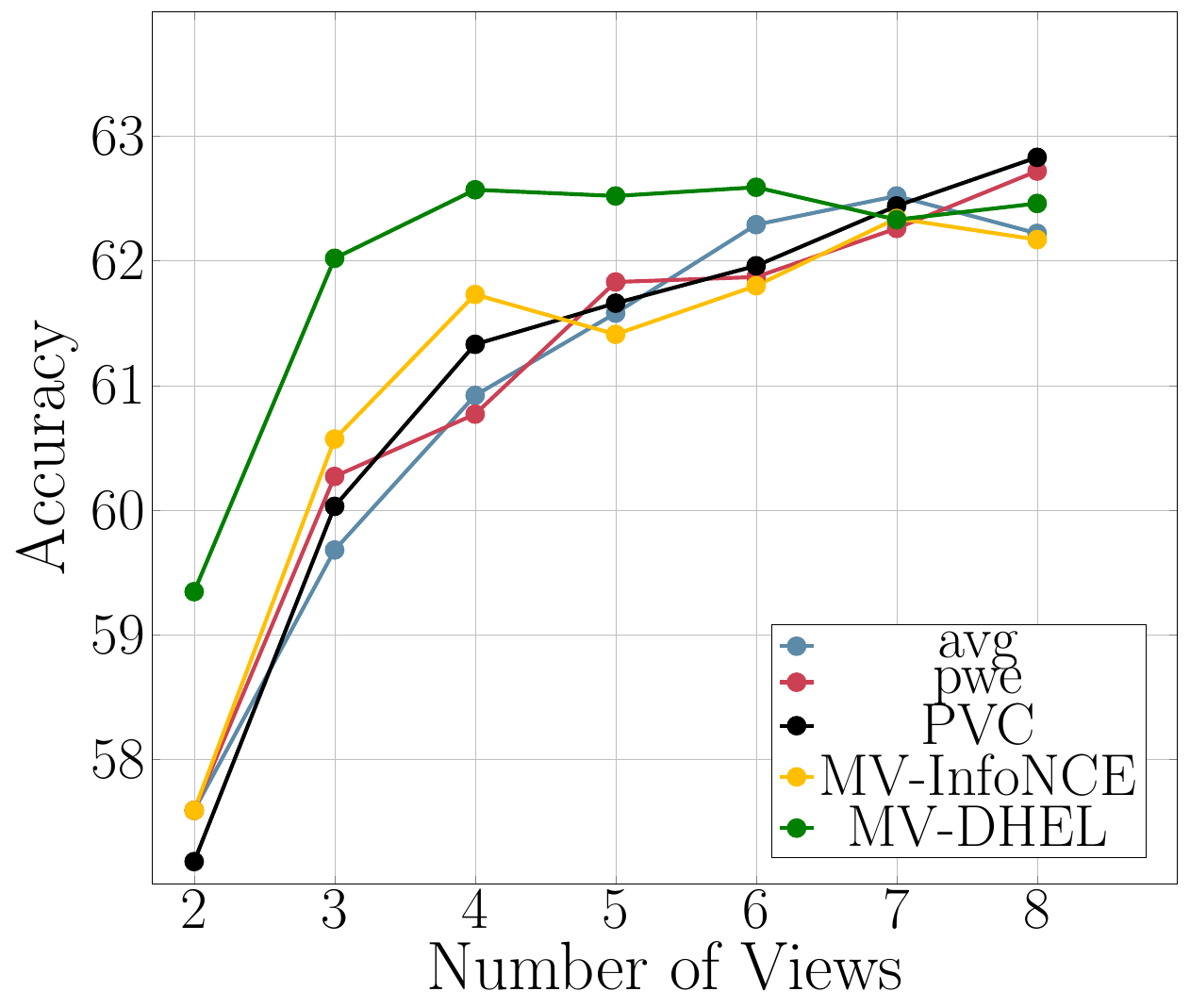}}
    \vspace{-2em}
    \caption*{(a) Downstream performance}
    \end{minipage}
    \begin{minipage}{0.23\textwidth}
    \resizebox{\textwidth}{!}{\includegraphics{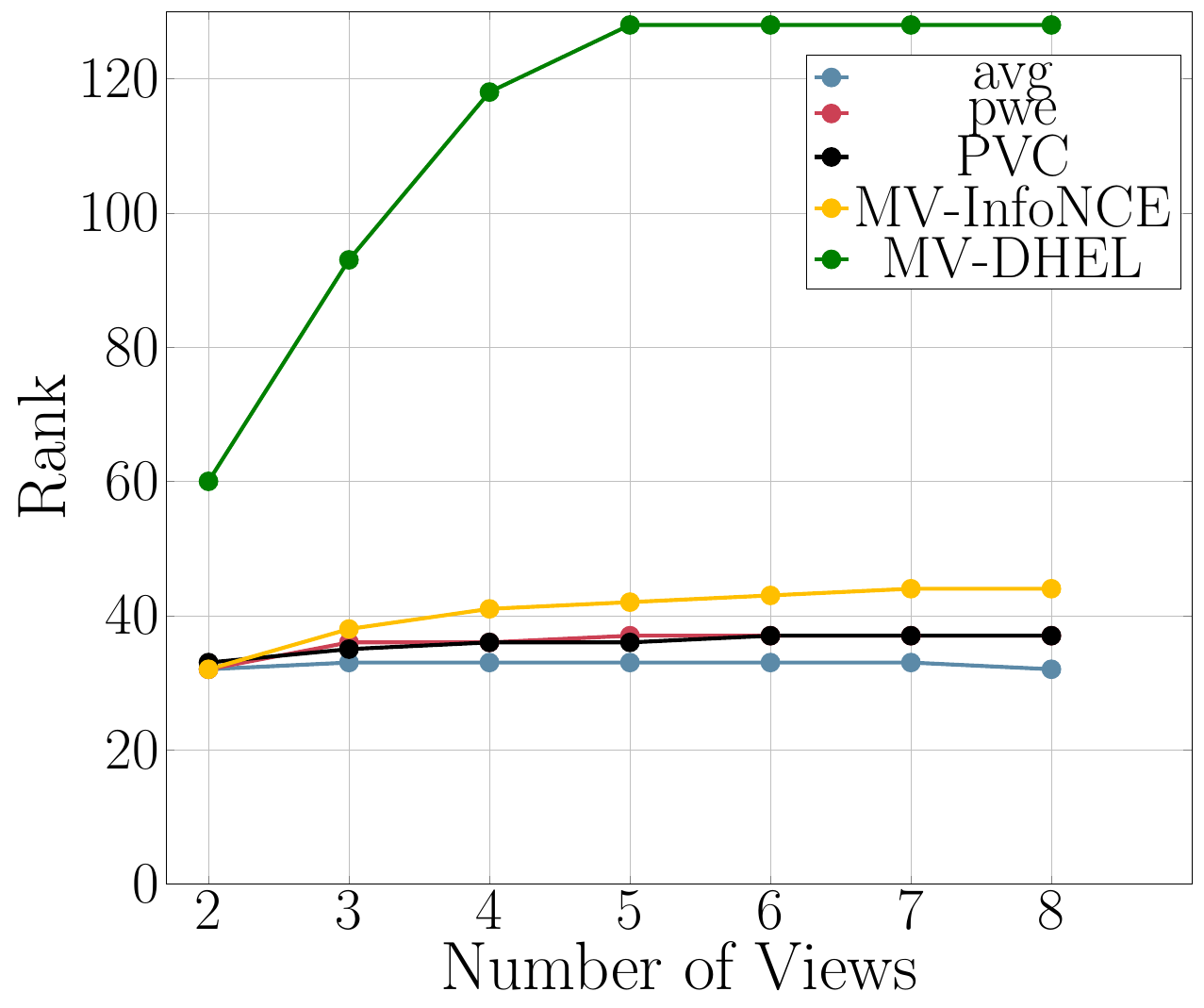}}
    \vspace{-2em}
    \caption*{(b) Rank}
    \end{minipage}
    \begin{minipage}{0.23\textwidth}
    \resizebox{\textwidth}{!}{\includegraphics{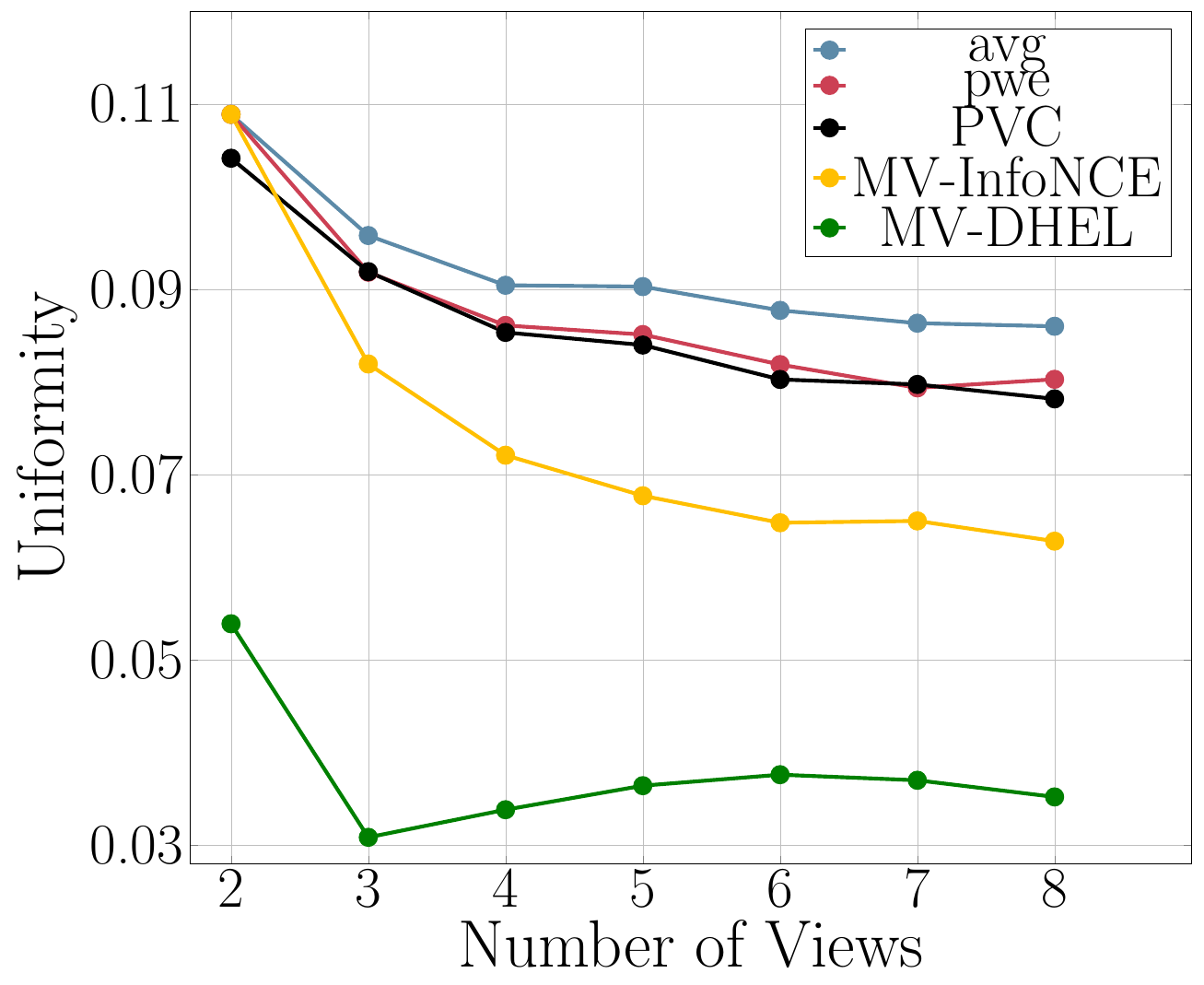}}
    \vspace{-2em}
    \caption*{(c) Uniformity}
    \end{minipage}
    \begin{minipage}{0.24\textwidth}
    \resizebox{\textwidth}{!}{\includegraphics{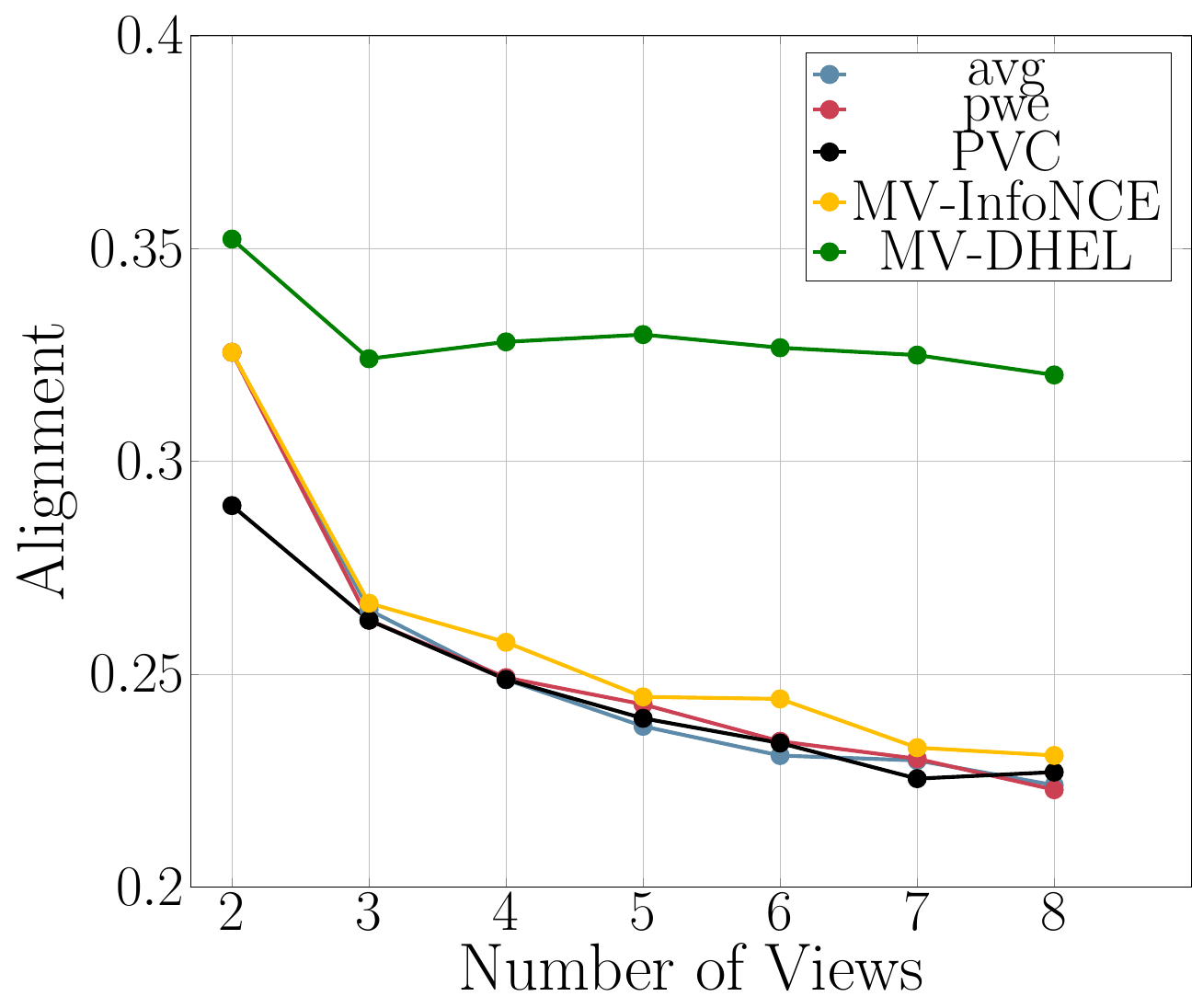}}
    \vspace{-2em}
    \caption*{(d) Alignment}
    \end{minipage}
    \caption{Properties vs view multiplicity calculated on CIFAR10 (top) \& CIFAR100 (bottom) dataset}\label{fig:properties_views}
    \vspace{-1em}
\end{figure*}

Here, we conduct a series of ablations on CIFAR10 and CIFAR100 with a batch size of 128, exploring various aspects of the proposed methods, including performance scaling, dimensionality collapse, optimization metrics, invariance to batch size and alternative multi-view batch samplings.
\subsubsection{Performance scaling with view multiplicity.} In \Cref{fig:properties_views}a, we show the performance scaling from 2 to 8 views for the CIFAR10 and CIFAR100 datasets. The results indicate that all methods improve as the number of views increases, with MVDHEL and MVInfoNCE consistently outperforming the other approaches across this broader range of views.

\subsubsection{Dimensionality collapse} In \Cref{fig:properties_views}b, we illustrate the rank of the matrix of learned embeddings by varying view multiplicity, which reflects the dimensionality utilised and the model's capacity for linear separation of data \cite{cover1965geometrical, garrido2023rankme}. Our results reveal that \textbf{MV-DHEL} (i) effectively \textbf{uses more dimensions as view multiplicity increases}, and (ii) at a sufficient number of views, it begins to \textbf{leverage the full 128-dimensional space}, mitigating dimensionality collapse. In contrast, competitors \textit{fail to leverage extra views to scale the dimensionality} of the learned embeddings.

\subsubsection{Optimisation} In \Cref{fig:properties_views}c and \ref{fig:properties_views}d, we present metrics that assess optimization quality. Specifically, we measure: (1) \textbf{Uniformity} \cite{wang2020understanding} using an improved metric from \cite{koromilas2024bridging} that, unlike the conventional uniformity metric \cite{wang2020understanding}, does not depend on a Gaussian kernel or require selecting a parameter 
t; and (2) \textbf{Alignment}, which estimates the expected distance between positive pair representations.

All baselines \textit{display similar optimization patterns} for alignment and uniformity, likely due to their common reliance on linear combinations of pairwise losses. In contrast, our methods create a \textit{distinct optimization landscape}, achieving \textit{significantly better uniformity}. This improvement scales with the number of views, with MV-DHEL distributing representations more evenly across dimensions, as seen in \Cref{fig:properties_views}b. 
MV-InfoNCE further achieves similar alignment to those of the baseline methods, while MV-DHEL exhibits worse alignment as is discussed in \cite{koromilas2024bridging}.

\begin{figure}[ht]
    \centering
   \begin{minipage}{0.21\textwidth}
    \resizebox{\textwidth}{!}{\includegraphics{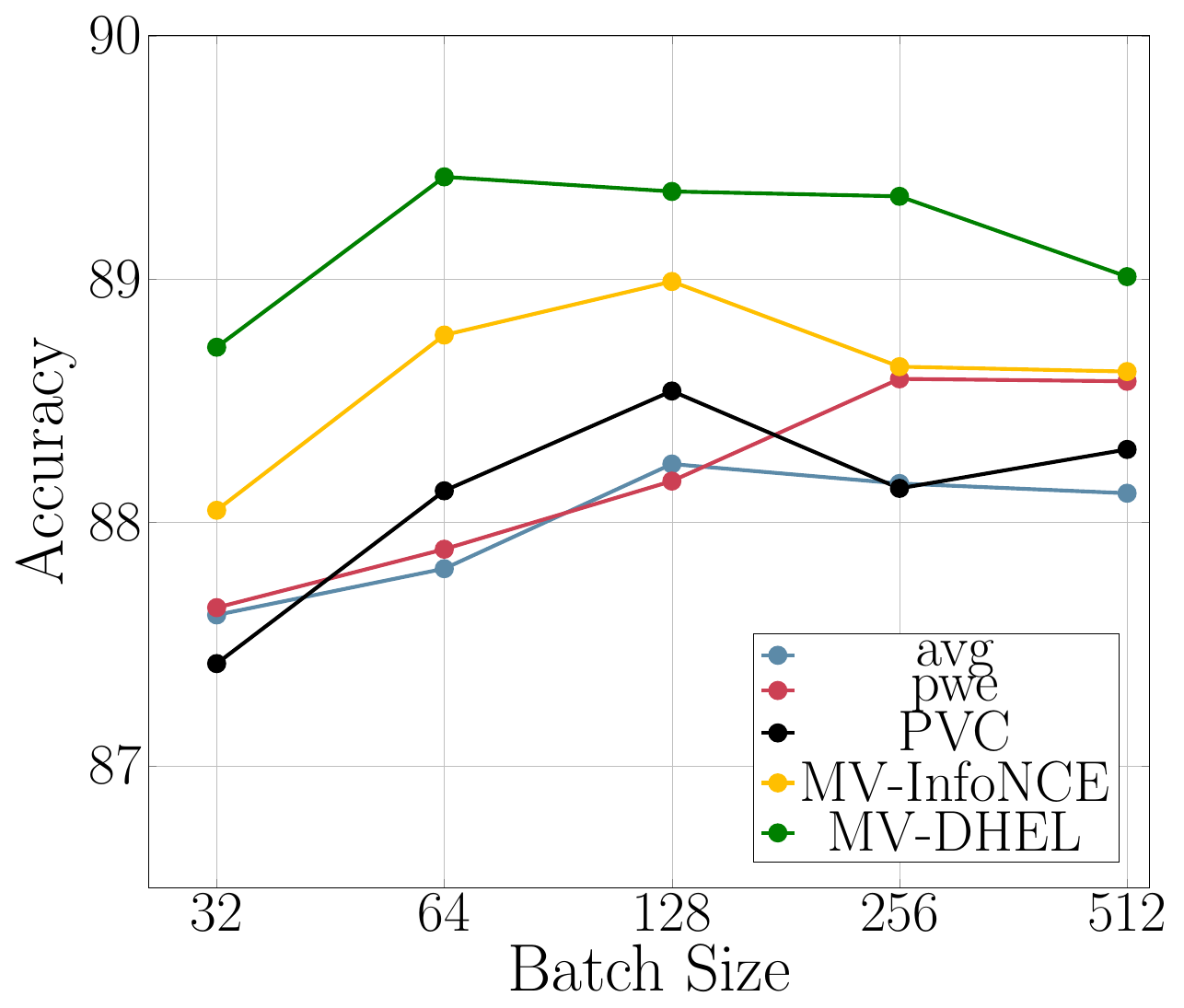}}
    \end{minipage}
    \begin{minipage}{0.21\textwidth}
    \resizebox{\textwidth}{!}{\includegraphics{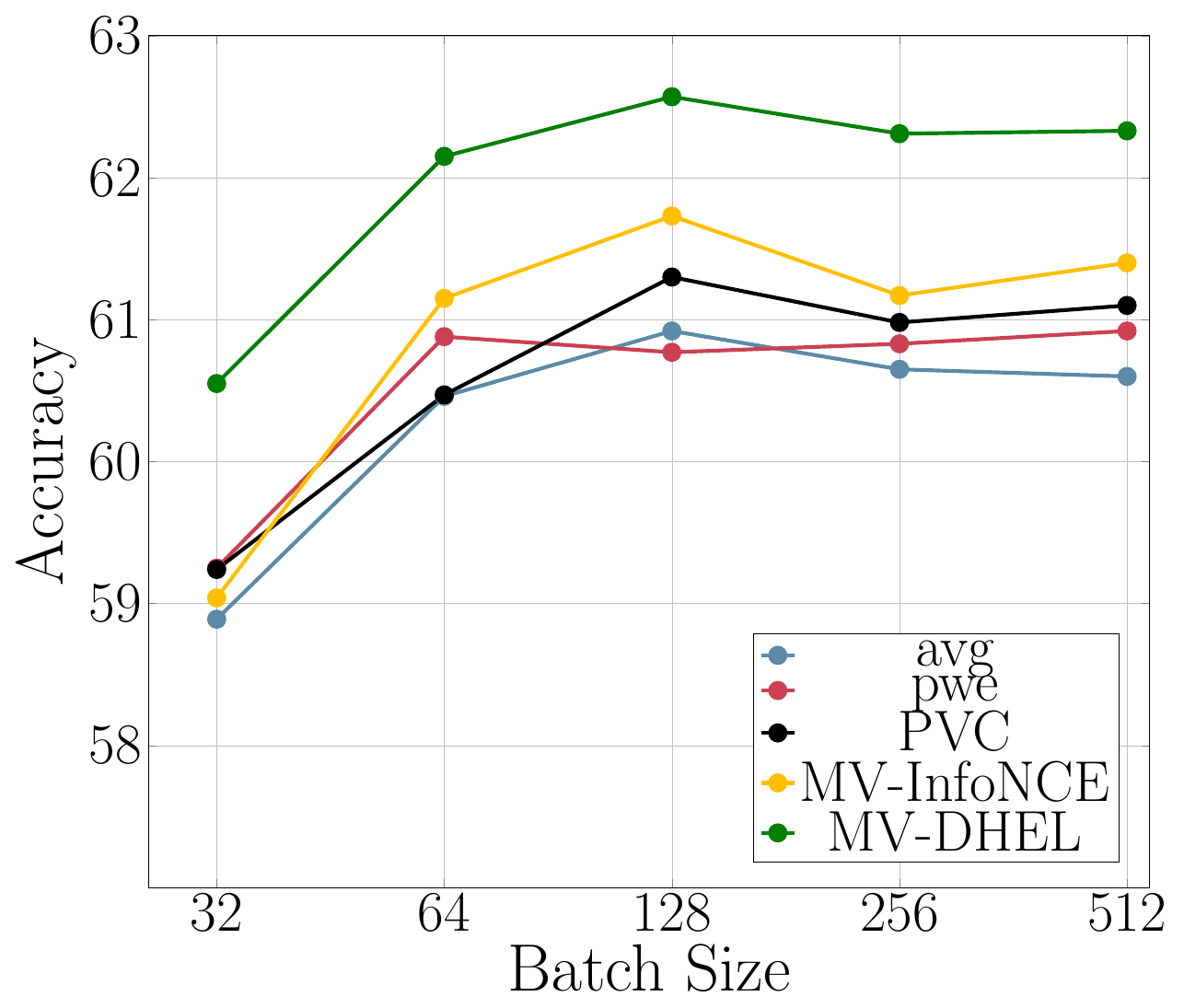}}
    \end{minipage} 
    \begin{minipage}{0.21\textwidth}
    \resizebox{\textwidth}{!}{\includegraphics{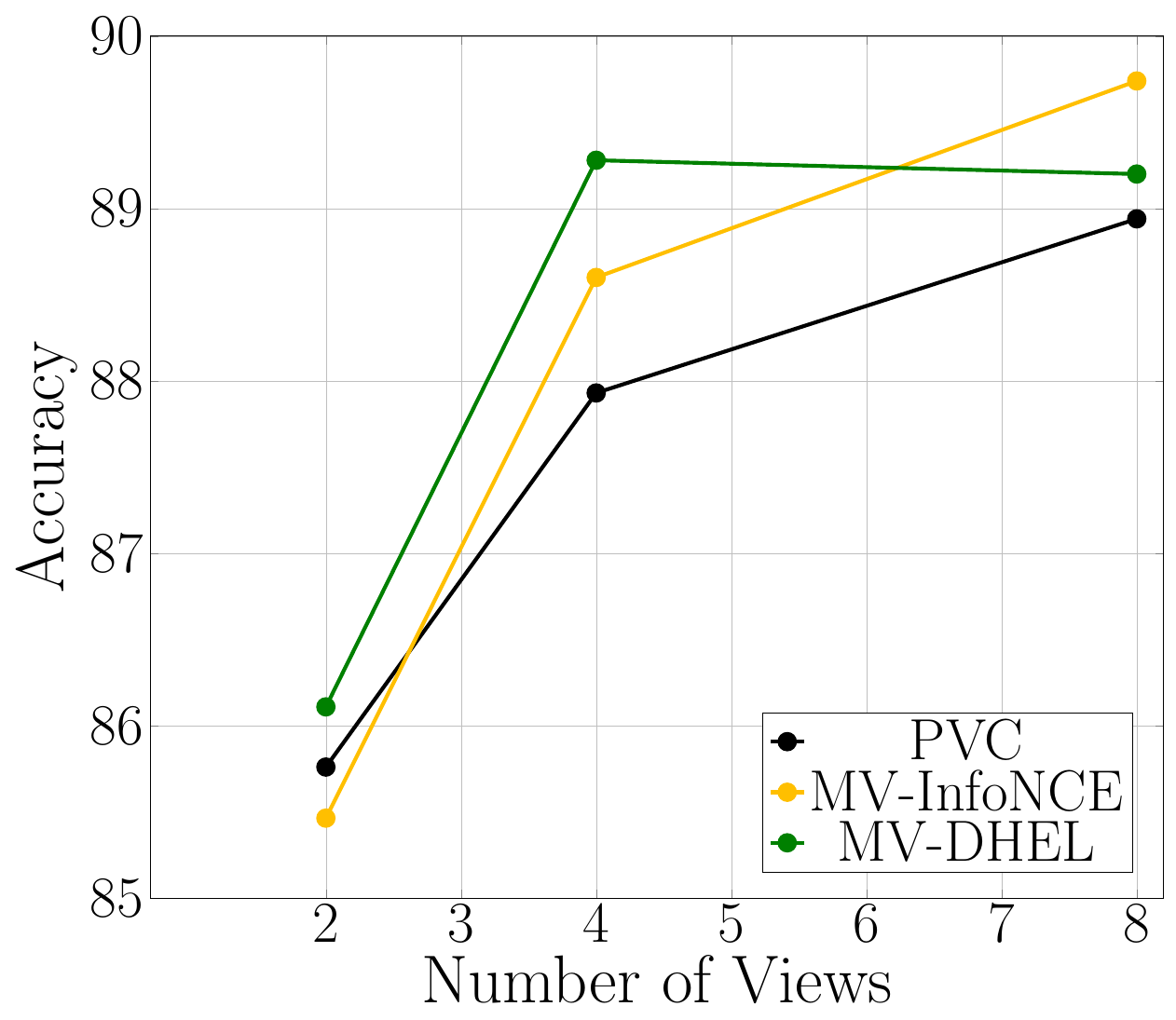}}
    \caption*{(a) CIFAR10}
    \end{minipage}
    \begin{minipage}{0.21\textwidth}
    \resizebox{\textwidth}{!}{\includegraphics{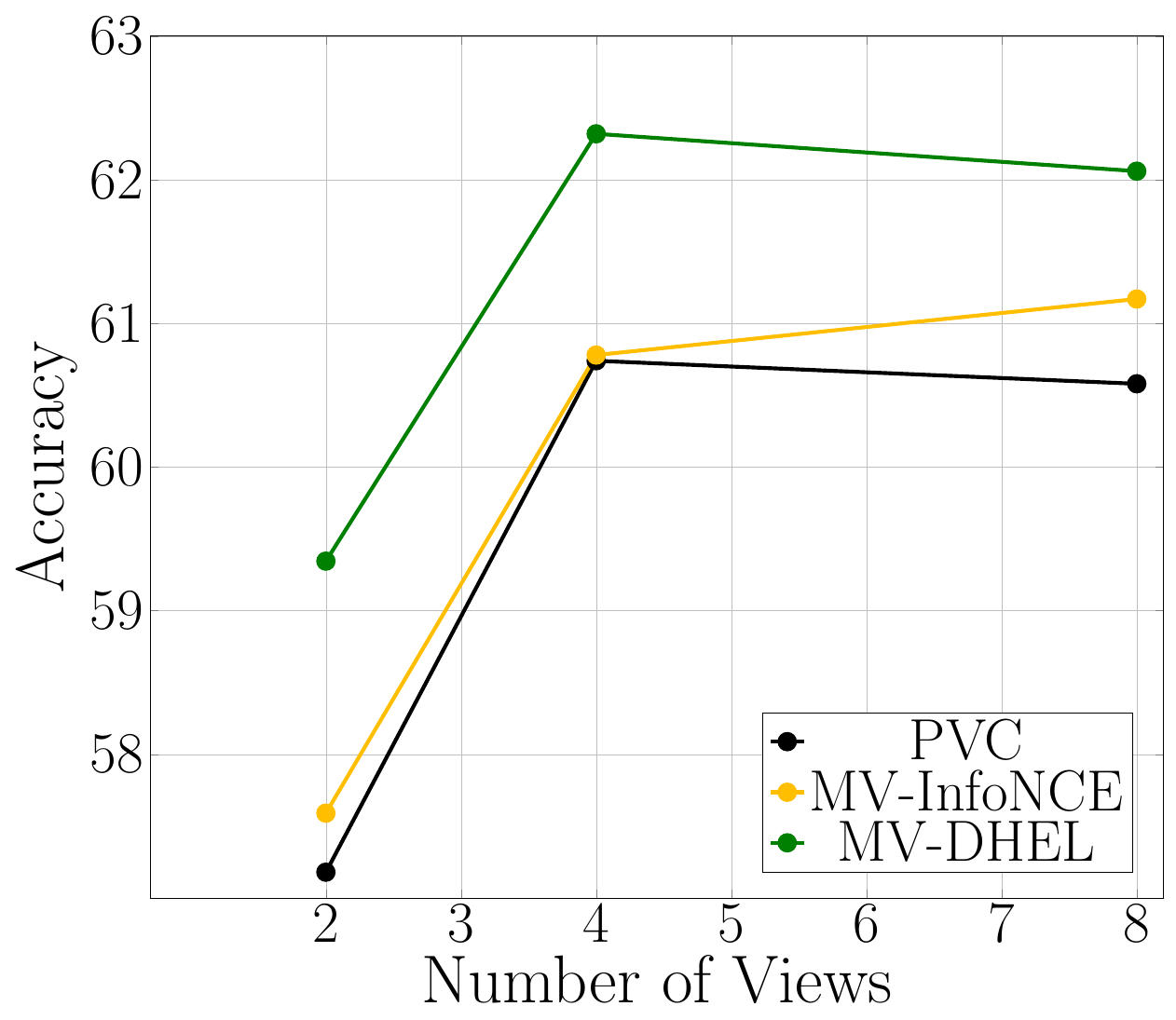}}
    \caption*{(b) CIFAR100}
    \end{minipage}
    \caption{Top: Performance vs batch size for 4 views; Bottom: Performance vs view multiplicity for fixed batch size.}\label{fig:batch}
\end{figure}

\subsubsection{Performance under varying batch sizes and fixed view multiplicity} In \Cref{fig:batch}(top), we illustrate the effect of varying batch sizes with a fixed view multiplicity of 4. The results show that pairwise aggregation methods are more sensitive to batch size, while MV-DHEL remains more stable across different batch sizes. This stability demonstrates that MV-DHEL is more effective in \textbf{leveraging view multiplicity to reduce the reliance on large batch sizes}.

\subsubsection{Performance under varying view multiplicity and fixed batch size}  While multiple views of each data point improve downstream performance and enhance properties, e.g. rank, it also significantly increases the actual batch size during the network’s forward pass, leading to higher memory demands. An alternative approach to accommodate more views is to \textit{keep the batch size fixed by reducing the number of unique data instances} \cite{fort2021drawing}. For example, instead of using 256 unique instances with a batch size of 256, one could use only 64 unique instances with 4 views per instance, resulting in an effective batch size of \(4 \times 64\). As seen in \Cref{fig:batch}(bottom), under a fixed batch size, only MV-DHEL reproduces the performance observed in \Cref{fig:properties_views}a, where increasing the number of views enhances performance. This ability to support multiple views without increasing memory usage stems from DHEL’s robustness to batch size variations \cite{koromilas2024bridging}.

\subsubsection{Memory overhead}
\Cref{fig:batch} (bottom) demonstrates that MV-DHEL scales to more views without incurring additional memory costs. \Cref{tab:memory_overhead} further confirms that MV-DHEL maintains stable performance with 4 views across different actual batch sizes on ImageNet-100, CIFAR-10, and CIFAR-100. These results reveal two key properties: invariance to actual batch size when using a fixed number of views, and performance improvement with increasing view multiplicity when using a fixed batch size. Together, they show that we can increase the number of views and benefit from the associated performance gains without requiring additional memory. This is primarily due to: (i) DHEL's robustness to batch size variations \cite{koromilas2024bridging}, and (ii) the positive impact of increased view multiplicity, which compensates for smaller batch sizes.

Using this trick for suitable views (N) and batch (M) maintains efficiency, e.g. InfoNCE (\( \mathcal{O}(M^2) \)) with M=1024, N=2 costs more than MV-DHEL (\( \mathcal{O}(N M^2) \)) with M=512, N=4.

\begin{table}[ht!]
\label{tab:memory_overhead}
\centering
\caption{4-view MV-DHEL performance for different number of unique instances per batch}
\begin{tabular}{|c|c|c|c|c|}
\hline
\textbf{Unique instances per batch} & \textbf{64} & \textbf{128} & \textbf{256} & \textbf{Deviation} \\ \hline
CIFAR10 & 89.49 & 89.52 & 89.47 & 0.05 \\ \hline
CIFAR100 & 62.43 & 62.78 & 62.73 & 0.35 \\ \hline
ImageNet-100 & 77.32 & 77.38 & 77.23 & 0.15 \\ \hline
\end{tabular}
\end{table}

\section{Conclusion}

We presented a principled approach to multi-view CL through two theoretically grounded objectives: MV-InfoNCE and MV-DHEL. Unlike current methods that process views through pair-wise loss aggregations, our framework enables interactions between all views simultaneously. Concretely, MV-InfoNCE generalises InfoNCE to handle multiple views, while MV-DHEL further addresses alignment-uniformity coupling. Our extensive experiments demonstrate key advantages: (i) improved downstream accuracy, (ii) better scalability as the number of views increases, and (iii) effective mitigation of dimensionality collapse when using MV-DHEL with several views.

\section*{Acknowledgments}
Panagiotis Koromilas was supported by the Hellenic Foundation for Research and Innovation (HFRI) under the 4th Call for HFRI PhD Fellowships (Fellowship Number:  10816). Giorgos Bouritsas and Yannis Panagakis were supported by the project MIS 5154714 of the National Recovery and Resilience Plan Greece 2.0 funded by the European Union under the NextGenerationEU Program. This research was partially supported by a grant from The Cyprus Institute on Cyclone.

\bibliographystyle{IEEEtran}
\bibliography{main}

\begin{thebibliography}{10}
\providecommand{\url}[1]{#1}
\csname url@samestyle\endcsname
\providecommand{\newblock}{\relax}
\providecommand{\bibinfo}[2]{#2}
\providecommand{\BIBentrySTDinterwordspacing}{\spaceskip=0pt\relax}
\providecommand{\BIBentryALTinterwordstretchfactor}{4}
\providecommand{\BIBentryALTinterwordspacing}{\spaceskip=\fontdimen2\font plus
\BIBentryALTinterwordstretchfactor\fontdimen3\font minus \fontdimen4\font\relax}
\providecommand{\BIBforeignlanguage}[2]{{%
\expandafter\ifx\csname l@#1\endcsname\relax
\typeout{** WARNING: IEEEtran.bst: No hyphenation pattern has been}%
\typeout{** loaded for the language `#1'. Using the pattern for}%
\typeout{** the default language instead.}%
\else
\language=\csname l@#1\endcsname
\fi
#2}}
\providecommand{\BIBdecl}{\relax}
\BIBdecl

\bibitem{wang2020understanding}
T.~Wang and P.~Isola, ``Understanding contrastive representation learning through alignment and uniformity on the hypersphere,'' in \emph{International Conference on Machine Learning (ICML)}.\hskip 1em plus 0.5em minus 0.4em\relax PMLR, 2020, pp. 9929--9939.

\bibitem{philip2019learning}
B.~Philip, H.~R. Devon, B.~William \emph{et~al.}, ``Learning representations by maximizing mutual information across views,'' \emph{Advances in neural information processing systems}, vol.~32, pp. 15\,535--15\,545, 2019.

\bibitem{fort2021drawing}
S.~Fort, A.~Brock, R.~Pascanu, S.~De, and S.~L. Smith, ``Drawing multiple augmentation samples per image during training efficiently decreases test error,'' \emph{arXiv preprint arXiv:2105.13343}, 2021.

\bibitem{hoffer2020augment}
E.~Hoffer, T.~Ben-Nun, I.~Hubara, N.~Giladi, T.~Hoefler, and D.~Soudry, ``Augment your batch: Improving generalization through instance repetition,'' in \emph{Proceedings of the IEEE/CVF Conference on Computer Vision and Pattern Recognition}, 2020, pp. 8129--8138.

\bibitem{lin2024good}
C.-H. Lin, C.~Kaushik, E.~L. Dyer, and V.~Muthukumar, ``The good, the bad and the ugly sides of data augmentation: An implicit spectral regularization perspective,'' \emph{Journal of Machine Learning Research}, vol.~25, no.~91, pp. 1--85, 2024.

\bibitem{caron2020unsupervised}
M.~Caron, I.~Misra, J.~Mairal, P.~Goyal, P.~Bojanowski, and A.~Joulin, ``Unsupervised learning of visual features by contrasting cluster assignments,'' \emph{Advances in neural information processing systems}, vol.~33, pp. 9912--9924, 2020.

\bibitem{caron2021emerging}
M.~Caron, H.~Touvron, I.~Misra, H.~J\'egou, J.~Mairal, P.~Bojanowski, and A.~Joulin, ``Emerging properties in self-supervised vision transformers,'' in \emph{Proceedings of the International Conference on Computer Vision (ICCV)}, 2021.

\bibitem{bardes2022vicregl}
A.~Bardes, J.~Ponce, and Y.~LeCun, ``Vicregl: Self-supervised learning of local visual features,'' \emph{Advances in Neural Information Processing Systems}, vol.~35, pp. 8799--8810, 2022.

\bibitem{tian2020contrastive}
Y.~Tian, D.~Krishnan, and P.~Isola, ``Contrastive multiview coding,'' in \emph{Computer Vision--ECCV 2020: 16th European Conference, Glasgow, UK, August 23--28, 2020, Proceedings, Part XI 16}.\hskip 1em plus 0.5em minus 0.4em\relax Springer, 2020, pp. 776--794.

\bibitem{shidani2024polyview}
\BIBentryALTinterwordspacing
A.~Shidani, R.~D. Hjelm, J.~Ramapuram, R.~Webb, E.~G. Dhekane, and D.~Busbridge, ``Poly-view contrastive learning,'' in \emph{The Twelfth International Conference on Learning Representations}, 2024. [Online]. Available: \url{https://openreview.net/forum?id=iHcTLIor0m}
\BIBentrySTDinterwordspacing

\bibitem{koromilas2024bridging}
\BIBentryALTinterwordspacing
P.~Koromilas, G.~Bouritsas, T.~Giannakopoulos, M.~Nicolaou, and Y.~Panagakis, ``Bridging mini-batch and asymptotic analysis in contrastive learning: From info{NCE} to kernel-based losses,'' in \emph{Forty-first International Conference on Machine Learning}, 2024. [Online]. Available: \url{https://openreview.net/forum?id=SvvvB5t5EW}
\BIBentrySTDinterwordspacing

\bibitem{jing2022understanding}
\BIBentryALTinterwordspacing
L.~Jing, P.~Vincent, Y.~LeCun, and Y.~Tian, ``Understanding dimensional collapse in contrastive self-supervised learning,'' in \emph{International Conference on Learning Representations}, 2022. [Online]. Available: \url{https://openreview.net/forum?id=YevsQ05DEN7}
\BIBentrySTDinterwordspacing

\bibitem{radford2021learning}
A.~Radford, J.~W. Kim, C.~Hallacy, A.~Ramesh, G.~Goh, S.~Agarwal, G.~Sastry, A.~Askell, P.~Mishkin, J.~Clark \emph{et~al.}, ``Learning transferable visual models from natural language supervision,'' in \emph{International conference on machine learning}.\hskip 1em plus 0.5em minus 0.4em\relax PmLR, 2021, pp. 8748--8763.

\bibitem{ruan2024tricolo}
Y.~Ruan, H.-H. Lee, Y.~Zhang, K.~Zhang, and A.~X. Chang, ``Tricolo: Trimodal contrastive loss for text to shape retrieval,'' in \emph{Proceedings of the IEEE/CVF Winter Conference on Applications of Computer Vision}, 2024, pp. 5815--5825.

\bibitem{liu2021contrastive}
Y.~Liu, Q.~Fan, S.~Zhang, H.~Dong, T.~Funkhouser, and L.~Yi, ``Contrastive multimodal fusion with tupleinfonce,'' in \emph{Proceedings of the IEEE/CVF International Conference on Computer Vision}, 2021, pp. 754--763.

\bibitem{sun2024contextual}
K.~Sun, Z.~Xie, M.~Ye, and H.~Zhang, ``Contextual augmented global contrast for multimodal intent recognition,'' in \emph{Proceedings of the IEEE/CVF Conference on Computer Vision and Pattern Recognition}, 2024, pp. 26\,963--26\,973.

\bibitem{chopra2005learning}
S.~Chopra, R.~Hadsell, and Y.~LeCun, ``Learning a similarity metric discriminatively, with application to face verification,'' in \emph{2005 IEEE computer society conference on computer vision and pattern recognition (CVPR'05)}, vol.~1.\hskip 1em plus 0.5em minus 0.4em\relax IEEE, 2005, pp. 539--546.

\bibitem{sohn2016improved}
K.~Sohn, ``Improved deep metric learning with multi-class n-pair loss objective,'' \emph{Advances in neural information processing systems}, vol.~29, 2016.

\bibitem{oord2018representation}
A.~v.~d. Oord, Y.~Li, and O.~Vinyals, ``Representation learning with contrastive predictive coding,'' \emph{arXiv preprint arXiv:1807.03748}, 2018.

\bibitem{chen2020simple}
T.~Chen, S.~Kornblith, M.~Norouzi, and G.~Hinton, ``A simple framework for contrastive learning of visual representations,'' in \emph{International conference on Machine Learning (ICML)}.\hskip 1em plus 0.5em minus 0.4em\relax PMLR, 2020, pp. 1597--1607.

\bibitem{dwibedi2021little}
D.~Dwibedi, Y.~Aytar, J.~Tompson, P.~Sermanet, and A.~Zisserman, ``With a little help from my friends: Nearest-neighbor contrastive learning of visual representations,'' in \emph{Proceedings of the IEEE/CVF International Conference on Computer Vision}, 2021, pp. 9588--9597.

\bibitem{yeh2022decoupled}
C.-H. Yeh, C.-Y. Hong, Y.-C. Hsu, T.-L. Liu, Y.~Chen, and Y.~LeCun, ``Decoupled contrastive learning,'' in \emph{European Conference on Computer Vision}.\hskip 1em plus 0.5em minus 0.4em\relax Springer, 2022, pp. 668--684.

\bibitem{he2020momentum}
K.~He, H.~Fan, Y.~Wu, S.~Xie, and R.~Girshick, ``Momentum contrast for unsupervised visual representation learning,'' in \emph{Proceedings of the IEEE/CVF Conference on Computer Vision and Pattern Recognition (CVPR)}, 2020, pp. 9729--9738.

\bibitem{RobinsonCSJ21}
J.~D. Robinson, C.~Chuang, S.~Sra, and S.~Jegelka, ``Contrastive learning with hard negative samples,'' in \emph{9th International Conference on Learning Representations, {ICLR} 2021, Virtual Event, Austria, May 3-7, 2021}.\hskip 1em plus 0.5em minus 0.4em\relax OpenReview.net, 2021.

\bibitem{hua2021feature}
T.~Hua, W.~Wang, Z.~Xue, S.~Ren, Y.~Wang, and H.~Zhao, ``On feature decorrelation in self-supervised learning,'' in \emph{Proceedings of the IEEE/CVF International Conference on Computer Vision}, 2021, pp. 9598--9608.

\bibitem{JingVLT22}
L.~Jing, P.~Vincent, Y.~LeCun, and Y.~Tian, ``Understanding dimensional collapse in contrastive self-supervised learning,'' in \emph{The Tenth International Conference on Learning Representations, {ICLR} 2022, Virtual Event, April 25-29, 2022}.\hskip 1em plus 0.5em minus 0.4em\relax OpenReview.net, 2022.

\bibitem{geiping2023how}
\BIBentryALTinterwordspacing
J.~Geiping, M.~Goldblum, G.~Somepalli, R.~Shwartz-Ziv, T.~Goldstein, and A.~G. Wilson, ``How much data are augmentations worth? an investigation into scaling laws, invariance, and implicit regularization,'' in \emph{The Eleventh International Conference on Learning Representations}, 2023. [Online]. Available: \url{https://openreview.net/forum?id=3aQs3MCSexD}
\BIBentrySTDinterwordspacing

\bibitem{bouchacourt2021grounding}
D.~Bouchacourt, M.~Ibrahim, and A.~Morcos, ``Grounding inductive biases in natural images: invariance stems from variations in data,'' \emph{Advances in Neural Information Processing Systems}, vol.~34, pp. 19\,566--19\,579, 2021.

\bibitem{tong2023empssl}
S.~Tong, Y.~Chen, Y.~Ma, and Y.~Lecun, ``Emp-ssl: Towards self-supervised learning in one training epoch,'' \emph{arXiv preprint arXiv:2304.03977}, 2023.

\bibitem{ermolov2021whitening}
A.~Ermolov, A.~Siarohin, E.~Sangineto, and N.~Sebe, ``Whitening for self-supervised representation learning,'' in \emph{International Conference on Machine Learning}.\hskip 1em plus 0.5em minus 0.4em\relax PMLR, 2021, pp. 3015--3024.

\bibitem{wang2024adaptive}
L.~Wang, P.~Koniusz, T.~Gedeon, and L.~Zheng, ``Adaptive multi-head contrastive learning,'' in \emph{European Conference on Computer Vision}.\hskip 1em plus 0.5em minus 0.4em\relax Springer, 2024, pp. 404--421.

\bibitem{li2023global}
J.~Li, B.~Liang, X.~Lu, M.~Li, G.~Lu, and Y.~Xu, ``From global to local: Multi-patch and multi-scale contrastive similarity learning for unsupervised defocus blur detection,'' \emph{IEEE Transactions on Image Processing}, vol.~32, pp. 1158--1169, 2023.

\bibitem{shah2023multi}
K.~Shah, A.~Shah, C.~P. Lau, C.~M. de~Melo, and R.~Chellappa, ``Multi-view action recognition using contrastive learning,'' in \emph{Proceedings of the ieee/cvf winter conference on applications of computer vision}, 2023, pp. 3381--3391.

\bibitem{xu2022multi}
J.~Xu, H.~Tang, Y.~Ren, L.~Peng, X.~Zhu, and L.~He, ``Multi-level feature learning for contrastive multi-view clustering,'' in \emph{Proceedings of the IEEE/CVF conference on computer vision and pattern recognition}, 2022, pp. 16\,051--16\,060.

\bibitem{piran2024contrasting}
\BIBentryALTinterwordspacing
Z.~Piran, M.~Klein, J.~Thornton, and marco cuturi, ``Contrasting multiple representations with the multi-marginal matching gap,'' in \emph{Forty-first International Conference on Machine Learning}, 2024. [Online]. Available: \url{https://openreview.net/forum?id=dV9B9qFeGi}
\BIBentrySTDinterwordspacing

\bibitem{grill2020bootstrap}
J.-B. Grill, F.~Strub, F.~Altch{\'e}, C.~Tallec, P.~Richemond, E.~Buchatskaya, C.~Doersch, B.~Avila~Pires, Z.~Guo, M.~Gheshlaghi~Azar \emph{et~al.}, ``Bootstrap your own latent-a new approach to self-supervised learning,'' \emph{Advances in neural information processing systems}, vol.~33, pp. 21\,271--21\,284, 2020.

\bibitem{pototzky2022fastsiam}
D.~Pototzky, A.~Sultan, and L.~Schmidt-Thieme, ``Fastsiam: Resource-efficient self-supervised learning on a single gpu,'' in \emph{DAGM German Conference on Pattern Recognition}.\hskip 1em plus 0.5em minus 0.4em\relax Springer, 2022, pp. 53--67.

\bibitem{wang2021unsupervised}
X.~Wang, Z.~Liu, and S.~X. Yu, ``Unsupervised feature learning by cross-level instance-group discrimination,'' in \emph{Proceedings of the IEEE/CVF conference on computer vision and pattern recognition}, 2021, pp. 12\,586--12\,595.

\bibitem{zhang2022dual}
C.~Zhang, K.~Zhang, T.~X. Pham, A.~Niu, Z.~Qiao, C.~D. Yoo, and I.~S. Kweon, ``Dual temperature helps contrastive learning without many negative samples: Towards understanding and simplifying moco,'' in \emph{Proceedings of the IEEE/CVF Conference on Computer Vision and Pattern Recognition}, 2022, pp. 14\,441--14\,450.

\bibitem{wu2018unsupervised}
Z.~Wu, Y.~Xiong, S.~X. Yu, and D.~Lin, ``Unsupervised feature learning via non-parametric instance discrimination,'' in \emph{Proceedings of the IEEE conference on computer vision and pattern recognition}, 2018, pp. 3733--3742.

\bibitem{liang2023multizoo}
P.~P. Liang, Y.~Lyu, X.~Fan, A.~Agarwal, Y.~Cheng, L.-P. Morency, and R.~Salakhutdinov, ``Multizoo \& multibench: A standardized toolkit for multimodal deep learning,'' \emph{Journal of Machine Learning Research}, vol.~24, pp. 1--7, 2023.

\bibitem{mao2022msena}
H.~Mao, Z.~Yuan, H.~Xu, W.~Yu, Y.~Liu, and K.~Gao, ``M-sena: An integrated platform for multimodal sentiment analysis,'' in \emph{Proceedings of the 60th Annual Meeting of the Association for Computational Linguistics: System Demonstrations}, 2022, pp. 204--213.

\bibitem{bagher-zadeh-etal-2018-multimodal}
\BIBentryALTinterwordspacing
A.~Bagher~Zadeh, P.~P. Liang, S.~Poria, E.~Cambria, and L.-P. Morency, ``Multimodal language analysis in the wild: {CMU}-{MOSEI} dataset and interpretable dynamic fusion graph,'' in \emph{Proceedings of the 56th Annual Meeting of the Association for Computational Linguistics (Volume 1: Long Papers)}, I.~Gurevych and Y.~Miyao, Eds.\hskip 1em plus 0.5em minus 0.4em\relax Melbourne, Australia: Association for Computational Linguistics, Jul. 2018, pp. 2236--2246. [Online]. Available: \url{https://aclanthology.org/P18-1208/}
\BIBentrySTDinterwordspacing

\bibitem{yu2020ch}
W.~Yu, H.~Xu, F.~Meng, Y.~Zhu, Y.~Ma, J.~Wu, J.~Zou, and K.~Yang, ``Ch-sims: A chinese multimodal sentiment analysis dataset with fine-grained annotation of modality,'' in \emph{Proceedings of the 58th annual meeting of the association for computational linguistics}, 2020, pp. 3718--3727.

\bibitem{cover1965geometrical}
T.~M. Cover, ``Geometrical and statistical properties of systems of linear inequalities with applications in pattern recognition,'' \emph{IEEE transactions on electronic computers}, no.~3, pp. 326--334, 1965.

\bibitem{garrido2023rankme}
Q.~Garrido, R.~Balestriero, L.~Najman, and Y.~Lecun, ``Rankme: Assessing the downstream performance of pretrained self-supervised representations by their rank,'' in \emph{International Conference on Machine Learning}.\hskip 1em plus 0.5em minus 0.4em\relax PMLR, 2023, pp. 10\,929--10\,974.

\bibitem{lian2024merbench}
Z.~Lian, L.~Sun, Y.~Ren, H.~Gu, H.~Sun, L.~Chen, B.~Liu, and J.~Tao, ``Merbench: A unified evaluation benchmark for multimodal emotion recognition,'' \emph{arXiv preprint arXiv:2401.03429}, 2024.

\bibitem{devlin2019bert}
J.~Devlin, M.-W. Chang, K.~Lee, and K.~Toutanova, ``{BERT}: Pre-training of deep bidirectional transformers for language understanding,'' in \emph{Proceedings of the 2019 Conference of the North {A}merican Chapter of the Association for Computational Linguistics: Human Language Technologies, Volume 1 (Long and Short Papers)}.\hskip 1em plus 0.5em minus 0.4em\relax Association for Computational Linguistics, Jun. 2019, pp. 4171--4186.

\bibitem{hsu2021hubert}
W.-N. Hsu, B.~Bolte, Y.-H.~H. Tsai, K.~Lakhotia, R.~Salakhutdinov, and A.~Mohamed, ``Hubert: Self-supervised speech representation learning by masked prediction of hidden units,'' \emph{IEEE/ACM Transactions on Audio, Speech, and Language Processing}, vol.~29, pp. 3451--3460, 2021.

\bibitem{dosovitskiy2021an}
\BIBentryALTinterwordspacing
A.~Dosovitskiy, L.~Beyer, A.~Kolesnikov, D.~Weissenborn, X.~Zhai, T.~Unterthiner, M.~Dehghani, M.~Minderer, G.~Heigold, S.~Gelly, J.~Uszkoreit, and N.~Houlsby, ``An image is worth 16x16 words: Transformers for image recognition at scale,'' in \emph{International Conference on Learning Representations}, 2021. [Online]. Available: \url{https://openreview.net/forum?id=YicbFdNTTy}
\BIBentrySTDinterwordspacing

\bibitem{loshchilov2018decoupled}
\BIBentryALTinterwordspacing
I.~Loshchilov and F.~Hutter, ``Decoupled weight decay regularization,'' in \emph{International Conference on Learning Representations}, 2019. [Online]. Available: \url{https://openreview.net/forum?id=Bkg6RiCqY7}
\BIBentrySTDinterwordspacing

\end{thebibliography}


{
\onecolumn
\appendix
\section*{Additional Preliminaries and Technical Details}
\label{sec:ad_notation}
\subsection{Kernels}
Minimizing the energy, ie the kernel potential $\sum_{\bx,\by} = K(\bx, \by)$, of a point configuration has shown to have direct connection to the optimisation of contrastive learning objectives \cite{wang2020understanding, koromilas2024bridging}. Kernels are of the form ${K(\bx, \by) = \kappa(\|\bx - \by\|^2)}$, with $\kappa: (0,4] \to \mathbb{R}$ and the limit $\underset{x \to 0^+}{\lim}\kappa(x)$ exists and is bounded, and $\gamma>0$ is a weighting coefficient.

Notable examples of kernels that obey the conditions that we encounter in this paper are the following: 
\begin{itemize}
    \item \textit{Gaussian}: ${\Gauss_t(\bx, \by) = e^{-t \|\bx - \by \|^2}}$ $= \gauss_t (\|\bx-\by\|^2)$, where $\gauss_t(x;t) = e^{-t x}$.
    \item \textit{Logarithmic}:
    ${\Logar_{s, \beta}(\bx, \by) = -\frac{1}{2}\log\left(s \|\bx - \by \|^{2} + \beta\right)}$$=\logar_{s, \beta} (\|\bx-\by\|^2)$, where $\logar_{s, \beta}(x) = -\frac{1}{2}\log \left( sx + \beta\right)$.
\end{itemize}

With elementary derivations, it is easy to see that $\gauss_t$ is strictly decreasing and convex for $t > 0$. For $s>-2$, $\riesz_s$ is strictly decreasing and strictly convex, while the same holds for $\logar_{s, \beta}$ when $s,\beta>0$. 

Additionally, for $t>0$, $\gauss_t$ is strictly completely monotone, while the same holds for $\logar_{s, \beta}$ for $s,\beta>0$.

\label{ssec:kernels}
\subsection{Mini-Batch Objectives}
\label{ssec:minibatch}

Here we re-write the PVC, MV-InfoNCE and MV-DHEL objectives in the InfoNCE form by applying the gaussian kernel on equations (\ref{eq:pvc}), (\ref{eq:mvinfonce_mb}), (\ref{eq:mvdhel_mb}).

\begin{equation}
\begin{split}
      \Lpvc(\tU) &= \frac{-1}{M (N-1)}\bigg( \sum_{l\in[N],\textcolor{ForestGreen}{l' \in[N] \setminus l} \atop i\in[M]}
      \log K(\tU_{i,l,:}, \tU_{i,\textcolor{ForestGreen}{l'},:}) + \sum_{l, l'\in[N]\setminus l
      \atop i\in[M]} \log\big(\sum_{\textcolor{BrickRed}{j \in [M]\setminus i \atop {m\in[N]}}}
    K(\tU_{i,l,:}, \tU_{\textcolor{BrickRed}{j,m},:}) + K(\tU_{i,l,:}, \tU_{{j,l'},:})\big)\bigg) \\
    &= \frac{-1}{M (N-1)} \sum_{l\in[N],\textcolor{ForestGreen}{l' \in[N] \setminus l} \atop i\in[M]} \log \left(\frac{ e^{\tU_{i,l,:}^{\top} \tU_{i,\textcolor{ForestGreen}{l'},:} / \tau}}{ e^{\tU_{i,l,:}^{\top} \tU_{i,\textcolor{ForestGreen}{l'},:} / \tau} + \sum_{\textcolor{BrickRed}{j \in [M]\setminus i \atop {m\in[N]}}} e^{\tU_{i,l,:}^{\top} \tU_{\textcolor{BrickRed}{j,m},:} / \tau}}\right)
\end{split}
\end{equation}

\begin{equation}
    \begin{split}
    \Lmvinfonce(\tU) &= 
\frac{1}{M} \sum_{i \in [M]} -\log{\sum_{l\in[N] \atop
 \textcolor{ForestGreen}{l' \in [N] \setminus l}} K(\tU_{i,l,:}^{\top} \tU_{i, \textcolor{ForestGreen}{l'},:})}
+\frac{1}{M} \sum_{i \in [M]}\log{\sum_{l \in [N] \atop {\textcolor{BrickRed}{j \in [M] \atop m \in [N]\setminus l}}}K(\tU_{i,l,:}^{\top} \tU_{\textcolor{BrickRed}{j,m},:})}\\
&= \frac{1}{M} \sum_{i=1}^M-\log \left(\frac{ \sum_{l\in[N] \atop
 \textcolor{ForestGreen}{l' \in [N] \setminus l}}^N e^{\tU_{i,l,:}^{\top} \tU_{i,\textcolor{ForestGreen}{l'},:}{/\tau}}}{ \sum_{l \in [N] \atop {\textcolor{BrickRed}{j \in [M] \atop m \in [N]\setminus l}}} e^{\tU_{i,l,:}^{\top} \tU_{\textcolor{BrickRed}{j,m,}:}{/\tau}}}\right)
\end{split}
\end{equation}

\begin{equation}
    \begin{split}
\Lmvdhel(\tU)
&= \frac{1}{M} \sum_{i \in [M]} -\log{\sum_{l\in[N] \atop
 \textcolor{ForestGreen}{l' \in [N] \setminus l}} K(\tU_{i,l,:}^{\top} \tU_{i,\textcolor{ForestGreen}{l'},:})} +\frac{1}{M} \sum_{l \in [N] \atop i \in [M]}\log{\sum_{\textcolor{BrickRed}{j \in [M]\setminus i}}K(\tU_{i,l,:}^{\top} \tU_{\textcolor{BrickRed}{j,l},:})}\\
 & = \frac{1}{M} \sum_{i=1}^M-\log \left(\frac{ \sum_{l\in[N] \atop
 \textcolor{ForestGreen}{l' \in [N] \setminus l}}^N e^{\tU_{i,l,:}^{\top} \tU_{i,\textcolor{ForestGreen}{l'},:}{/\tau}}}{ \prod_{l \in [N]} \sum_{{\textcolor{BrickRed}{j \in [M]}}} e^{\tU_{i,l,:}^{\top} \tU_{\textcolor{BrickRed}{j,m,}:}{/\tau}}}\right)
    \end{split}
\end{equation}

Here, in order to properly compare our functions, we further define two more multi-view contrastive losses two different losses which are obtained by using the alignment to \cref{eq:mv_align_out}, the uniformity to \cref{eq:mv_unif_out} and the negative index sets based on the corresponding sets of MV-InfoNCE and MV-DHEL respectively.

\begin{equation}
\begin{split}
L_{\textnormal{MV-CL1}}(\tU) &= 
\frac{1}{NM} \sum_{l\in[N] \atop i \in [M]} -\log{\sum_{\textcolor{ForestGreen}{l' \in [N] \setminus l}} K(\tU_{i,l,:}^{\top} \tU_{i, \textcolor{ForestGreen}{l'},:})}
+\frac{1}{NM} \sum_{l\in[N] \atop i \in [M]}\log{\sum_{{\textcolor{BrickRed}{j \in [M] \atop m \in [N]\setminus l}}}K(\tU_{i,l,:}^{\top} \tU_{\textcolor{BrickRed}{j,l},:})}\\  
&= \frac{1}{M} \sum_{l\in[N] \atop i \in [M]}-\log \left(\frac{ \sum_{ \textcolor{ForestGreen}{l' \in [N] \setminus l}}^N e^{\tU_{i,l,:}^{\top} \tU_{i,\textcolor{ForestGreen}{l'},:}{/\tau}}}{ \sum_{{\textcolor{BrickRed}{j \in [M] \atop m \in [N]\setminus l}}} e^{\tU_{i,l,:}^{\top} \tU_{\textcolor{BrickRed}{j,m,}:}{/\tau}}}\right)
\end{split}
\label{eq:mvinfonce_mb_theory}
\end{equation}

\begin{equation}
    \begin{split}
L_{\textnormal{MV-CL2}}(\tU)
&= \frac{1}{NM} \sum_{l\in[N] \atop i \in [M]} -\log{\sum_{
 \textcolor{ForestGreen}{l' \in [N] \setminus l}} K(\tU_{i,l,:}^{\top} \tU_{i,\textcolor{ForestGreen}{l'},:})}
+\frac{1}{NM} \sum_{l \in [N] \atop i \in [M]}\log{\sum_{\textcolor{BrickRed}{j \in [M]\setminus i}}K(\tU_{i,l,:}^{\top} \tU_{\textcolor{BrickRed}{j,l},:})}.\\
 & = \frac{1}{M} \sum_{ l \in [N] \atop i \in [M]}-\log \left(\frac{ \sum_{
 \textcolor{ForestGreen}{l' \in [N] \setminus l}}^N e^{\tU_{i,l,:}^{\top} \tU_{i,\textcolor{ForestGreen}{l'},:}{/\tau}}}{ \sum_{{\textcolor{BrickRed}{j \in [M]}}} e^{\tU_{i,l,:}^{\top} \tU_{\textcolor{BrickRed}{j,m,}:}{/\tau}}}\right)
    \end{split}
    \label{eq:mvdhel_mb_theory}
\end{equation}

\subsection{Expectations of Mini-Batch Objectives}
\label{ssec:expectations}

We will denote the pushforward measures induced by $f$ (encoder) with $f_\#\pdata$. Additionally, we denote with $p_{\text{trans}}$ the distribution of a datapoint sampled by $p_{\text{init}}$ and then transformed by a single transformation sampled by $\cT$.
By sampling a tensor of $M$ datapoints of \( N \) positive views from the pushforward measure induced by \( f \) on the $N$-view distribution, \ie
 \( \bU_{j} = (\bu_1, \dots, \bu_N) \overset{\text{i.i.d}}{\sim} \funsimple_\#\pdata \), $j \in [M]$. The  the mini-batch objectives are estimators of the following expectations:

 \begin{equation}
\begin{split}
&E_{\text{PVC}} = 
\underset{
\bU_{j}\overset{\text{i.i.d}}{\sim} \funsimple_\#\pdata^{M}}{\E}\left[ \sum_{l\in[N] \atop {l' \in[N] \setminus l}} - \log \left(\frac{ e^{\tU_{1,l,:}^{\top} \tU_{1,l',:} / \tau}}{ e^{\tU_{1,l,:}^{\top} \tU_{1,l',:} / \tau} + \sum_{{j \in [M]\setminus i \atop {m\in[N]}}} e^{\tU_{1,l,:}^{\top} \tU_{j,m,:} / \tau}}\right)\right]
\end{split}
\label{eq:pvc_expectation}
\end{equation}

 \begin{equation}
\begin{split}
&E_{\text{MV-InfoNCE}} = 
\underset{
\bU_{j}\overset{\text{i.i.d}}{\sim} \funsimple_\#\pdata^{M}}{\E}\left[-\log \left(\frac{\sum_{l \in [N], \atop l' \in [N]\setminus l} e^{\bU_{1,l}^{\top} \bU_{1,l'}{/\tau}}}{\sum_{l \in [N] \atop {j\in [M]\atop m \in [N]\setminus l}} e^{\bU_{1,l}^{\top} \bU_{j, m}{/\tau}}}\right)\right]
\end{split}
\label{eq:mvinfonce_expectation_supp}
\end{equation}
\begin{equation}
\begin{split}
&E_{\text{MV-DHEL}} =  \underset{
\bU_{j}\overset{\text{i.i.d}}{\sim} \funsimple_\#\pdata^{M}}{\E}\left[-\log \left(\frac{\sum_{l \in [N], \atop l' \in [N]\setminus l}  e^{\bU_{1,l}^{\top} \bU_{1,l'}{/\tau}}}{
\prod_{l \in [N]}\sum_{{j \in  [M-1]}} e^{\bU_{1,l}^{\top} \bU_{j, l}{/\tau}}}\right)\right]
\end{split}
\label{eq:mvdhel_expectation_supp}
\end{equation}

 \section*{Proof of Theorem 4.1}
\label{sec:proof}

\noindent\textbf{Theorem 4.1.} The expectations of the following batch-level contrastive loss functions: $\Linfoncea(\cdot, \cdot)$, $\Lmvinfonce(\cdot, \cdot)$, $\Lmvdhel(\cdot, \cdot)$, have the \textbf{same asymptotic behaviour} when normalized by appropriate normalizing constants.

\begin{proof}

For convenience we repeat the expected values that are going to analyse subtracting the appropriate normalising constants:
\begin{equation}
\begin{split}
&E_1 = 
\underset{
\bU_{j}\overset{\text{i.i.d}}{\sim} \funsimple_\#\pdata^{M}}{\E}\left[-\log \left(\frac{\sum_{l \in [N], \atop l' \in [N]\setminus l} e^{\bU_{1,l}^{\top} \bU_{1,l'}{/\tau}}}{\sum_{l \in [N] \atop {j\in [M]\atop m \in [N]\setminus l}} e^{\bU_{1,l}^{\top} \bU_{j, m}{/\tau}}}\right)\right] - \log (M-1) + \log(N(N-1))
\end{split}
\label{eq:mvinfonce_expectation_supp}
\end{equation}
\begin{equation}
\begin{split}
&E_2 =  \underset{
\bU_{j}\overset{\text{i.i.d}}{\sim} \funsimple_\#\pdata^{M}}{\E}\left[-\log \left(\frac{\sum_{l \in [N], \atop l' \in [N]\setminus l}  e^{\bU_{1,l}^{\top} \bU_{1,l'}{/\tau}}}{
\prod_{l \in [N]}\sum_{{j \in  [M-1]}} e^{\bU_{1,l}^{\top} \bU_{j, l}{/\tau}}}\right)\right] -\log(M-1) + \log(N(N-1))
\end{split}
\label{eq:mvdhel_expectation_supp}
\end{equation}

Following, set:
\begin{equation}
    \begin{split}
        A &=  -\underset{(\bU_{1,1}, ...\bU_{1,N}) \sim \funsimple_\#\pdata}{\E}\left[\log \left({\sum_{l\in[N] \atop l'\in [N] \setminus l} e^{\bU_{1,l}^{\top} \bU_{1,l'}{/\tau}}}\right)\right] + \log(N(N-1))\\
        B &= \underset{
        \bU_{j}\overset{\text{i.i.d}}{\sim} \funsimple_\#\pdata^{M}}{\E}\left[\log \left({\sum_{l \in [N] \atop {j\in [M]\atop m \in [N]\setminus l}} e^{\bU_{1,l}^{\top} \bU_{j, m}{/\tau}}}\right)\right] - \log(M-1)
 \\
        \Gamma &= \underset{
        \bU_{j}\overset{\text{i.i.d}}{\sim} \funsimple_\#\pdata^{M}}{\E}\left[\log \left({
        \prod_{l \in [N]}\sum_{{j \in  [M-1]}} e^{\bU_{1,l}^{\top} \bU_{j, l}{/\tau}}}\right)\right] + N\log(M-1)
        ,
    \end{split}
\end{equation}
such that $E_1 = A + B$ and $E_2 = A + \Gamma$. Now we expand each term as follows:

\underline{\noindent\textbf{Regarding A}}: Using Jensen's Inequality 
we get:
\begin{equation*}
    \begin{split}
    -A  &= 
    \underset{(\bU_{1,1}, ...\bU_{1,N}) \sim \funsimple_\#\pdata}{\E}\left[\log \left(\frac{\sum_{l\in[N] \atop l'\in [N] \setminus l} e^{\bU_{1,l}^{\top} \bU_{1,l'}{/\tau}}}{N(N-1)}\right)\right]\\
    &\leq
    \underset{(\bU_{1,1}, ...\bU_{1,N}) \sim \funsimple_\#\pdata}{\E}\left[ \left(\frac{\sum_{l\in[N] \atop l'\in [N] \setminus l} \log e^{\bU_{1,l}^{\top} \bU_{1,l'}{/\tau}}}{N(N-1)}\right)\right] \\
    &=   \underset{(\bU_{1,1}, ...\bU_{1,N}) \sim \funsimple_\#\pdata}{\E}\left[ \left(\frac{\sum_{l\in[N] \atop l'\in [N] \setminus l} 
    \bU_{1,l}^{\top} \bU_{1,l'}{/\tau}}{N(N-1)}\right)\right]\\
    &\leq   \underset{(\bU_{1,1}, ...\bU_{1,N}) \sim \funsimple_\#\pdata}{\E}\left[ \left(\frac{\sum_{l\in[N] \atop l'\in [N] \setminus l} 
    1 {/\tau}}{N(N-1)}\right)\right]\\
    &={1/\tau},
\end{split}
\end{equation*}
where in the last step we used the fact that inner products are maximised when the angle between the two vectors is zero, and since all vectors have unit norm then $\bU_{1,l}^{\top} \bU_{1,l'} \leq 1$. Therefore, the maximisation of the last term happens when $\bU_{1,l} = \bU_{1,l'} , \forall l \neq l' \in \{1, N\}$. Now, observe that in this case, Jensen's holds with equality since all summands are equal to $e^{1/\tau}$. Therefore, maximisation of $-A - \log\big(N(N-1)\big)$ happens when there exists an encoder $f$ such that whenever we sample from the pushforward $f_\#\pdata$ we obtain $\bU_{1,l} = \bU_{1,l'}, \forall l \neq l' \in \{1, N\}$, \ie, when there is perfect alignment between $\bU_{1,l}$ and $\bU_{1,l'}$ for all $l, l'$ \ie across views.

\underline{\textbf{Regarding B:}}

For fixed $\bU_{1,1}, ..., \bU_{1,N}$, dividing by $M-1$, and due to the law of the large numbers we have:

\begin{equation*}
    \begin{split}
        &\lim_{M \to \infty} \frac{1}{M-1}\left(\sum_{l\in[N]}\sum_{m \in [N]\setminus l}\sum_{j\in[M]\setminus 1} e^{\bU_{1,l}^{\top} \bU_{j, m}{/\tau}} + \sum_{l\in[N]}\sum_{m \in [N]\setminus l}  e^{\bU_{1,l}^{\top} \bU_{1, m}{/\tau}}\right)\\
        & = \underset{(\bu_1, ...\bu_n) \sim \pdata}{\E}\left[\sum_{l\in[N] \atop m \in [N]\setminus l}  e^{\bU_{1,l}^{\top} \bu_{m}{/\tau}}\right]\\
        & = \sum_{m \in [N]\setminus l}\underset{{y \sim p_{init}} \atop {(T_m) \sim \cT^N}}{\E}\left[\sum_{l\in[N]} e^{\bU_{1,l}^{\top} f(T_m(y)){/\tau}}\right]\\
        & = (N-1) \cdot \underset{{y \sim p_{init}} \atop {(T_m) \sim \cT^N}}{\E}\left[\sum_{l\in[N]} e^{\bU_{1,l}^{\top} f(T_m(y)){/\tau}}\right]\\
        & = (N-1) \cdot \underset{\bu \sim p_{trans}}{\E}\left[\sum_{l\in[N]} e^{\bU_{1,l}^{\top} \bu{/\tau}}\right]
    \end{split}
\end{equation*}

Now, the same limit holds for the $\log$ due to the Continuous Mapping theorem, thus:

\begin{equation*}
    \lim_{M \to \infty} \log \left({\sum_{l \in [N] \atop {j\in [M]\atop m \in [N]\setminus l}} e^{\bU_{1,l}^{\top} \bU_{j, m}{/\tau}}}\right) = \log \left( (N-1) \cdot \underset{\bu \sim p_{trans}}{\E}\left[\sum_{l\in[N]} e^{\bU_{1,l}^{\top} \bu{/\tau}}\right]\right)
\end{equation*}

Due to the Dominated Convergence Theorem:
\begin{equation*}
    \begin{split}
        \lim_{M \to \infty} B &= \underset{
        \bU_{j}\overset{\text{i.i.d}}{\sim} \funsimple_\#\pdata^{M}}{\E}\left[\lim_{M \to \infty} \log \left(\frac{1}{M-1}{\sum_{l \in [N] \atop {j\in [M]\atop m \in [N]\setminus l}} e^{\bU_{1,l}^{\top} \bU_{j, m}{/\tau}}} \right)\right]\\
        & = \underset{(\bU_{1,1}, ...\bU_{1,N}) \sim \funsimple_\#\pdata }{\E}\left[\log \left( (N-1) \cdot \underset{\bu \sim p_{trans}}{\E}\left[\sum_{l=1}^{N} e^{\bU_{1,l}^{\top} \bu{/\tau}}\right]\right)\right] \\
        & = \underset{(\bU_{1,1}, ...\bU_{1,N}) \sim \funsimple_\#\pdata }{\E}\left[\log \left( (N-1) \cdot \sum_{l=1}^{N} \underset{\bu \sim p_{trans}}{\E}\left[ e^{\bU_{1,l}^{\top} \bu{/\tau}}\right]\right)\right] \\
        & = \underset{{x \sim p_{init}} \atop (T_l) \sim \cT^N }{\E}\left[\log \left( (N-1) \cdot \sum_{l=1}^{N} \underset{\bu \sim p_{trans}}{\E}\left[ e^{f(T_l(x))^{\top} \bu{/\tau}}\right]\right)\right]\\
        & = \underset{{\bv \sim p_{trans}}}{\E}\left[\log \left( N(N-1) \cdot \underset{\bu \sim p_{trans}}{\E}\left[ e^{\bv^{\top} \bu{/\tau}}\right]\right)\right]\\
        & = \underset{{\bv \sim p_{trans}}}{\E}\left[\log \left( \underset{\bu \sim p_{trans}}{\E}\left[ e^{\bv^{\top} \bu{/\tau}}\right]\right)\right] + \log N(N-1)
    \end{split}
\end{equation*}

which, based on result 2 of Theorem 1 in \cite{wang2020understanding}, if perfectly uniform encoders exist, they form the exact minimizers of $\lim_{M \to \infty}B$.

\underline{\noindent\textbf{Regarding $\Gamma$}}: We deconstruct it via our sampling mechanism as follows:

\begin{equation}
    \begin{split}
        \Gamma + N\log(M-1)= & \underset{
        \bU_{j}\overset{\text{i.i.d}}{\sim} \funsimple_\#\pdata^{M}}{\E}\left[\log \left({
        \prod_{l \in [N]}\sum_{{j \in  [M-1]}} e^{\bU_{1,l}^{\top} \bU_{j, l}{/\tau}}}\right)\right] \\
        = & \underset{\bx \sim p_{init} \atop (\by_1, \dots, \by_{M-1}) \sim p_{init}^{M-1}}{\E}
        \underset{(T_{l,j}) \sim \cT^{NM}}{\E}
        \left[ \sum_{l=1}^{N} 
        \log \left(\sum_{j=1}^{M-1} e^{f\big(T_{l,M}(\bx)\big)^{\top}f\big(T_{l,j}(\by_j)\big){/\tau}}\right)
        \right]\\
        = & \underset{\bx \sim p_{init} \atop (\by_1, \dots, \by_{M-1}) \sim p_{init}^{M-1}}{\E}
        \left[ \sum_{l=1}^{N} \underset{(T_{l,j}) \sim \cT^{MN}}{\E}\left[\log \left(\sum_{j=1}^{M-1} e^{f\big(T_{l,M}(\bx)\big)^{\top}f\big(T_{l,j}(\by_j)\big){/\tau}}\right)\right]\right]\\
        = & \underset{\bx \sim p_{init} \atop (\by_1, \dots, \by_{M-1}) \sim p_{init}^{M-1}}{\E}
        \left[ \sum_{l=1}^{N} \underset{(T_{j}) \sim \cT^{M}}{\E}\left[\log \left(\sum_{j=1}^{M-1} e^{f\big(T_{M}(\bx)\big)^{\top}f\big(T_{j}(\by_j)\big){/\tau}}\right)\right]\right]\\
        = & \underset{\bx \sim p_{init} \atop (\by_1, \dots, \by_{M-1}) \sim p_{init}^{M-1}}{\E}
        \left[ N \underset{(T_{j}) \sim \cT^{M}}{\E}\left[\log \left(\sum_{j=1}^{M-1} e^{f\big(T_{M}(\bx)\big)^{\top}f\big(T_{j}(\by_j)\big){/\tau}}\right)\right]\right]\\
        = &  N \underset{\bv \sim \funsimple_\# {p_{\text{trans}}} \atop \bu_j\overset{\text{i.i.d}}{\sim} \funsimple_\#{p_{\text{trans}}}^{M-1}}{\E}\left[\log \left(\sum_{j=1}^{M-1} e^{\bv^{\top} \bu_j{/\tau}}\right)\right],
    \end{split}
\end{equation}

where ${p_{\text{trans}}}$ is the distribution of a single-view datatapoint transformed by a randomly sampled transformation. Now, for a fixed $\bv$ due to the strong law of large numbers, it holds that:

\begin{equation*}
    \lim_{M \to \infty} \frac{1}{M - 1}\sum_{j=1}^{M-1} e^{\bv^{\top} \bu_j{/\tau}} = \E_{u \sim p_{\text{trans}}}\left[ e^{\bv^{\top} \bu{/\tau}} \right]
\end{equation*}

Now using the same steps as in the proof of Theorem 1 in \cite{wang2020understanding} it follows that if perfectly uniform encoders exist they form the exact minimizers of $\Gamma$, when subtracting a normalisation constant M-1. Briefly, the same limit holds for the log (continuous function) of the above quantities due to the Continuous Mapping Theorem, and therefore when taking the limit of each loss variant (after first subtracting the normalisation constant  M - 1), since the quantities inside the expectation are bounded, we can invoke the Dominated Convergence Theorem and switch the limit with the expectation, thus arriving at the desideratum.

\underline{\textbf{Overall:}}

Based on results for $A$, $B$ and $\Gamma$, both $\Lmvinfonce$ and $\Lmvdhel$ have the same asymptotic behaviour as $\Linfoncea$ from \cite{wang2020understanding} when normalized by appropriate normalizing constants.
\end{proof}

\section*{Experimental Details}
\label{sec:impl_details}

In the following section, we provide a detailed description of the experimental setup.

\subsection{Detailed Sampling Process.}\label{ssec:sampling}

We collect multi-view datapoints as follows: First, we sample a data point $\bx_{\text{init}} \in \cX$ from the \textit{initial distribution} $p_{\text{init}}$ on $\cX$ (\ie\ the one from which we sample the data points in our dataset) and subsequently we independently sample $N$ \textit{transformation operators} ${T_i: \cX \to \cX}$ from a known distribution $p_{T}$ on a space of available transformations $\cT$. \textit{The transformation operators encode the symmetries of the data, \ie\ it is expected that the downstream tasks will be invariant to them.} The input distribution $p$ is defined as the distribution of the 
tuples of the form $[\bx_1; \dots ; \bx_N] = \left(T_1\left(\bx_{\text{init}}\right),..., T_N\left(\bx_{\text{init}}\right)\right)$.
and the p.d.f. is given by 
\begin{equation}
    \begin{split}
        &p(\bx_1,..., \bx_K)
        = \int_{T_1,...,T_K \in \cT, x_{\text{init}}\in \cX}p_{\text{init}}(x_{\text{init}}) p_T(T_1)\cdot...\cdot p_T(T_K) dx_{\text{init}}dT_1\cdot...\cdot dT_K
    \end{split}
\end{equation}, 
where $ x_1 = T_1(x_{\text{init}}),...,x_k = T_1K(x_{\text{init}})$.

The transformations are implemented as a series of resizing, cropping, horizontal flipping, color jittering, random grayscale conversion and gaussian blur.

\subsection{Implementation Details}\label{ssec:impl_details}
\paragraph{Code} The implementation of the experimental pipeline (networks, augmentations, training, evaluation functions etc) was based on \url{https://github.com/AndrewAtanov/simclr-pytorch.git} where we used our sampling process to sample N-view data and implemented all the reposrted loss functions. Our implementation can be found at \href{https://github.com/pakoromilas/Multi-View-CL.git}{https://github.com/pakoromilas/Multi-View-CL.git}.

\paragraph{CIFAR10 and CIFAR100} ResNet-18 is employed as the encoder architecture for CIFAR10 and CIFAR100 datasets. Training spans 200 epochs with the SGD optimizer and the cosine annealing learning rate schedule, using a base learning rate of (batch size) / 256. Augmentations include resizing, cropping, horizontal flipping, color jittering, and random grayscale conversion.  Linear evaluation is conducted by training a single linear layer on the learned embeddings, with an additional 200 epochs using SGD and a learning rate of 0.1.  We set the batch size to 256 and temperature to 0.5.

\paragraph{ImageNet-100} ResNet-50 is employed as the encoder architecture for ImageNet-100. Training spans 200 epochs with the SGD optimizer and the cosine annealing learning rate schedule, using a base learning rate of 1.4 * (batch size) / 256. We use the same augmentations as in the above datasets and extend them to include gaussian blur. Linear evaluation is conducted by training a single linear layer on the learned embeddings, with an additional 200 epochs using SGD and a learning rate of 0.5. We set the batch size to 256 and temperature to 0.5.

\paragraph{ImageNet1K} ResNet-50 is employed as the encoder architecture for ImageNet1K. Training spans 100 epochs with the LARS optimizer and the cosine annealing learning rate schedule, using a base learning rate of 0.3 * (batch size) / 256. We use the same augmentations as in ImageNet-100. Linear evaluation is conducted by training a single linear layer on the learned embeddings, with an additional 100 epochs using SGD and a learning rate of 1.6. We set the batch size to 512 and temperature to 0.1. We achieve to properly reproduce the results of \cite{chen2020simple} under the same batch size and number of epochs.

\paragraph{Augmentations} In the augmentation pipeline we apply each trasnformation with the same probability presented in \cite{chen2020simple} for all experiments.

\paragraph{CH-SIMS and CMU-MOSEI} A three layer transformer encoder is employed for each individual modality (unimodal encoder), followed by a late fusion concatenation operation, and a linear projection. This architecture is employed as the multimodal encoder architecture. The unimodal encoders are applied on top of extracted features following~\cite{lian2024merbench}. We employ BERT\footnote{\url{https://huggingface.co/google-bert/bert-base-chinese}}~\cite{devlin2019bert} for text, HuBERT\footnote{\url{https://huggingface.co/TencentGameMate/chinese-hubert-large}}~\cite{hsu2021hubert} for audio, and CLIP-ViT\footnote{\url{https://huggingface.co/openai/clip-vit-large-patch14}}~\cite{radford2021learning, dosovitskiy2021an} for visual components respectively. Contrastive training spans 200 epochs with learning rates tuned in the range $\{$$1e$-$5$, $5e$-$5$, $1e$-$4$$\}$ for each contrastive objective independently, and cosine annealing learning rate scheduler. All modalities are projected into a common space and treated as different views. Supervised training follows standard Multimodal Sentiment Analysis (MSA) literature~\cite{mao2022msena}, and trains a linear projection layer on the concatenated multimodal representation (vanilla late fusion) for 100 epochs. Training involves minimizing a regression loss, as MSA models predict continuous sentiment polarity values (in the range [-1,1] for our case). We utilize Mean Absolute Error (MAE) as the loss objective for CH-SIMS. All optimization processes employ the AdamW~\cite{loshchilov2018decoupled} optimizer for network weight updates. Contrastive training is repeated across three randomly selected seeds and we also fit three independent linear layers in the supervised tuning setup. We report the mean average across these experiments in this work, consistent with established practices in MSA literature~\cite{mao2022msena}.

\subsection{Comparison to other Multi-View losses}\label{sec:further_res}

In this section we compare the performance of the proposed MV-InfoNCE \Cref{eq:mvinfonce_mb}) and MV-DHEL (\Cref{eq:mvdhel_mb}) as presented in \Cref{sec:method} to other possible multi-view extensions MV-CL1 (\Cref{eq:mvinfonce_mb_theory}) and MV-CL2 (\Cref{eq:mvdhel_mb_theory}).
\begin{table}[ht]
\centering
\renewcommand{\arraystretch}{1.3} 
\begin{tabular}{|c|cc|cc|cc|cc|}
\hline
\textbf{\# Views} & \multicolumn{2}{c|}{\textbf{MV-CL1}} & \multicolumn{2}{c|}{\textbf{MV-InfoNCE}} & \multicolumn{2}{c|}{\textbf{MV-CL2}} & \multicolumn{2}{c|}{\textbf{MV-DHEL}} \\ 
 & \multicolumn{2}{c|}{\cref{eq:mvinfonce_mb_theory}} & \multicolumn{2}{c|}{\cref{eq:mvinfonce_mb}} & \multicolumn{2}{c|}{\cref{eq:mvdhel_mb_theory}} & \multicolumn{2}{c|}{\cref{eq:mvdhel_mb}} \\ \hline
 & \textbf{Acc.} & \textbf{Diff} & \textbf{Acc.} & \textbf{Diff} & \textbf{Acc.} & \textbf{Diff} & \textbf{Acc.} & \textbf{Diff} \\ \hline
2 & 72.2 & -- & 72.2 & -- & 73.3 & -- & 73.3 & -- \\
3 & 74.6 & +2.4 & 75.2 & +3.0 & 76.7 & +3.4 & \textbf{77.1} & \textbf{+3.8} \\
4 & 74.8 & +0.2 & 75.8 & +0.6 & 76.8 & +0.1 & {\color{ForestGreen}\textbf{77.2}} & +0.1 \\ \hline
\end{tabular}
\caption{Performance comparison of multi-view contrastive objectives on ImageNet-100. Bold values indicate the highest accuracy per view and the best improvement (Diff) from the previous view. MV-DHEL consistently outperforms other methods across all view configurations.}
\label{tab:mv_methods_comparison}
\end{table}

\begin{table}[ht]
\centering
\begin{tabular}{lcc}
\toprule
\multicolumn{3}{c}{\textbf{CH-SIMS}} \\ 
\textbf{Method} & \textbf{Accuracy}($\uparrow$) & \textbf{MAE}($\downarrow$) \\
\midrule
MV-CL1 (Eq. \ref{eq:mvinfonce_mb_theory})    &    75.19 & 0.434    \\ 
MV-InfoNCE (Eq. \ref{eq:mvinfonce_mb})    &    76.56 & 0.421    \\ 
MV-CL2 (Eq. \ref{eq:mvdhel_mb_theory})       & 79.08 & 0.394          \\
MV-DHEL (Eq. \ref{eq:mvdhel_mb})       & \efthygeo{\textbf{79.38}} & \efthygeo{\textbf{0.392}}          \\
\bottomrule
\end{tabular}%
\caption{Performance comparison for different possible multi-view loss extensions for the CH-SIMS multimodal dataset.}
\label{tab:results_mm_supp}
\end{table}

In \Cref{tab:mv_methods_comparison} and \Cref{tab:results_mm_supp}, we observe that MV-InfoNCE outperforms MV-CL1, which has the same negative index set, while MV-DHEL also performs better than MV-CL2, which shares the same negative index set, on both the demanding ImageNet-100 dataset and in the multimodal setup. We argue that this performance degradation originates from both MV-CL1 and MV-CL2 violating \textcolor{NavyBlue}{P3} by using multiple terms per data point.
}

\end{document}